\documentclass[11pt]{article}
\usepackage{blindtext}
\usepackage{jmlr2e}
\usepackage[english]{babel}
\usepackage{mathtools,mathrsfs}
\usepackage{algorithm}
\usepackage[noend]{algpseudocode}
\usepackage{listings, enumitem}
\usepackage{graphicx}
\usepackage[inkscapeformat=png]{svg}
\usepackage{subcaption}
\usepackage{tikz}
\usepackage[nomessages]{fp}
\usepackage{pgfplots}
\usepackage{booktabs}
\usepackage{siunitx}

\pgfplotsset{compat=1.11}
\pgfplotsset{
	colormap={parula}{
	rgb255=(53,42,135)
	rgb255=(15,92,221)
	rgb255=(18,125,216)
	rgb255=(7,156,207)
	rgb255=(21,177,180)
	rgb255=(89,189,140)
	rgb255=(165,190,107)
	rgb255=(225,185,82)
	rgb255=(252,206,46)
	rgb255=(249,251,14)}
}
\usepackage{hyperref}

\newcommand{\dx}{\,\mathrm{d}}

\DeclareMathOperator{\R}{{\mathbb{R}}}

\renewcommand{\vec}[1]{\mathbf{#1}}
\newcommand{\mat}[1]{\mathbf{#1}}

\DeclareMathOperator*{\argmin}{arg\,min} %handles subscripts like \lim
\DeclareMathOperator{\prox}{prox} %handle normal subscripts

\newcommand{\TV}[1]{\mathrm{TV}^{(#1)}}

\renewcommand{\vec}[1]{\mathbf{#1}}

\DeclareMathOperator{\Lip}{Lip}

\newcommand{\BV}[1]{\mathrm{BV}^{(#1)}}

\ShortHeadings{Lipschitz Networks with Learnable Activation Functions}{Ducotterd, Goujon, Bohra, Perdios, Neumayer, and Unser}
\firstpageno{1}

\begin{document}

\title{Improving Lipschitz-Constrained Neural Networks by Learning Activation Functions}

\author{\name Stanislas Ducotterd \email                       stanislas.ducotterd@epfl.ch \\ 
        \name Alexis Goujon \email alexis.goujon@epfl.ch \\ 
        \name Pakshal Bohra \email
        pakshal.bohra@epfl.ch \\
        \name Dimitris Perdios \email
        dimitris.perdios@epfl.ch \\
        \name Sebastian Neumayer \email
        sebastian.neumayer@epfl.ch \\
        \name Michael Unser \email
        michael.unser@epfl.ch \\
        \addr Biomedical Imaging Group,\\
        \'Ecole polytechnique f\'ed\'erale de Lausanne (EPFL),\\
        CH-1015 Lausanne, Switzerland}

\editor{}

\maketitle

\begin{abstract}
    Lipschitz-constrained neural networks have several advantages over unconstrained ones and can be applied to a variety of problems, making them a topic of attention in the deep learning community.
    Unfortunately, it has been shown both theoretically and empirically that they perform poorly when equipped with ReLU activation functions. By contrast, neural networks with learnable 1-Lipschitz linear splines are known to be more expressive. In this paper, we show that such networks correspond to global optima of a constrained functional optimization problem that consists of the training of a neural network composed of 1-Lipschitz linear layers and 1-Lipschitz freeform activation functions with second-order total-variation regularization. Further, we propose an efficient method to train these neural networks. 
    Our numerical experiments show that our trained networks compare favorably with existing 1-Lipschitz neural architectures.
\end{abstract}

\section{Introduction}
Lipschitz-constrained neural networks limit the maximum deviation of the output in response to a change of the input.
This property allows them to generalize well~\citep{10.5555/1005332.1005357, spectrally_norm, 10.5555/3295222.3295344, sokolic2017robust}, to be robust against adversarial attacks \citep{TSS2018, engstrom2019exploring, HN20, Pauli2022}, and to be more interpretable \citep{ross_improving_2017, DBLP:conf/iclr/TsiprasSETM19}.
They also appear in the training of Wasserstein generative adversarial networks (GAN) \citep{pmlr-v70-arjovsky17a}.
Finally, such networks can be inserted in iterative plug-and-play (PnP) algorithms to solve inverse problems, with the guarantee that the algorithm converges.
For successful applications of PnP algorithms in image reconstruction, we refer to \cite{SVW2016,  MMHC2017, ryu2019plug, hertrich_convolutional_2021}.

Unfortunately, the computation of the Lipschitz constant of a neural network is NP-hard, as shown in \cite{VS2018}.
Rather than prescribing the exact Lipschitz constant of a network, one usually either penalizes large Lipschitz constants through regularization, or constrains the Lipschitz constant of each linear layer and each activation function.
Regularization approaches \citep{cisse2017parseval, GAAD2017, BRRS2021} penalize the Lipschitz constant via a regularizer in the training loss.
They maintain good empirical performance, but do not offer a direct control of the Lipschitz constant of the network.

\paragraph{1-Lip Architectures}
In the constrained design, which is the one that will be considered here, one fixes the Lipschitz constant of each layer and of each activation function to one, resulting in what we refer to as \emph{1-Lip neural networks}.
There are several ways to impose constraints on the linear layers.
The most popular one is spectral normalization \citep{MKKY2018}, where the $\ell_2$ operator norm of each weight matrix is set to one. The required spectral norms are computed via power iterations.
To take this idea even further, \cite{ALG19} have restricted the weight matrices to be orthonormal in fully connected layers.

The use of rectified linear-unit (ReLU) activation functions in that setting, however, appears to be overly constraining: it has been shown that 1-Lip ReLU networks cannot even represent simple functions such as the absolute value function, both under 2-norm \citep{ALG19} and $\infty$-norm \citep{HCC2018} constraints on the linear layers.
This observation justifies the development of new activation functions specifically tailored to 1-Lip architectures.
Currently, the most popular one is GroupSort (GS), proposed by \cite{ALG19}, where the pre-activations are split into groups that are sorted in ascending order.
This results in a multivariate and gradient-norm-preserving (GNP) activation function.
The authors provide empirical evidence that GS outperforms ReLU on Wasserstein-$1$ distance estimation, robust classification, and function fitting under Lipschitz constraints.

We pursue an alternative way to boost the performance of 1-Lip neural networks.
Our motivation stems from several recent theoretical results in favor of linear-spline activation functions \citep{neumayer2022approximation}.
Notably, the authors prove that $1$-Lipschitz linear splines with three adjustable linear regions are capable of achieving the optimal expressivity among all 1-Lip networks with component-wise activation functions.
As those splines are unknown and potentially different for each neuron, we must learn them.

So far, there is no efficient implementation of 1-Lipschitz learnable linear-splines (LLS).
Fortunately, we can build upon existing frameworks for learning unconstrained linear-splines \citep{agostinelli2014learning, jin_deep_2016, BCGA2020}.
In this setting, \cite{Unser2019} has proven that neural networks with such activation functions are solutions of a functional optimization problem that consists of the training of a neural network with freeform activation functions whose second-order total variation is penalized.

% In this paper, we propose an alternative way to boost the performance of 1-Lip neural networks via the use of 1-Lipschitz learnable-linear-spline (LLS) activation functions.
% The observation that ReLU is the simplest instance of a nontrivial linear spline has inspired researchers to develop frameworks for the learning of more sophisticated linear-spline activation functions \citep{agostinelli2014learning, he_delving_2015, jin_deep_2016}.
% Moreover, \cite{Unser2019} has proven that neural networks with LLS activation functions are solutions of a functional optimization problem that consists of the training of a neural network with freeform activation functions whose second-order total variation is penalized.
% This has motivated the development of the deep spline neural network (DSNN) framework \citep{BCGA2020} for the efficient learning of such linear-spline activation functions \textcolor{red}{talk about knot advantage}.
% Unlike other approaches, DSNNs originate from a global functional optimization problem, which offers control over the number of effective linear regions of the activation functions. \textcolor{red}{why DSNN}
% Several theoretical results in favor of 1-Lip DSNNs have been established in \cite{neumayer2022approximation}, where the authors prove that LLSs with three linear regions already achieve the optimal expressivity among all 1-Lip networks with component-wise activation functions. 

\paragraph{Contribution}We extend the work of \cite{BCGA2020} to the Lipschitz-constrained setting.
In their experimental comparison, it was found to be the most efficient and stable LLS framework.
Since those parametric activation functions are a priori not $1$-Lipschitz, it is necessary to adapt the theory to this new setting and to develop computational tools to control the Lipschitz constant.
Here, our contributions are threefold.\\[-.5em]

\noindent \textbf{1.\ Theory:} We show that 1-Lip LLS networks correspond to the global optima of a {\em constrained functional-optimization problem}. The latter consists of the training of a neural network composed of 1-Lipschitz linear layers and 1-Lipschitz freeform activation functions with second-order total-variation regularization.
In particular, we prove that the solution of this problem always exists.
In effect, the 1-Lip constraint ensures stability, while the second-order total-variation regularization favors configurations with few linear regions.\\[-.5em]

\noindent \textbf{2.\ Implementation:} We formulate the training as an {\em unconstrained optimization problem} by representing the LLSs in a B-spline basis and by incorporating two novel modules.
\begin{enumerate}
    \item 
    An efficient method to explicitly control the Lipschitz constant of each LLS, which we call SplineProj.

    \item A normalization module that modulates the scale of each LLS without changing their Lipschitz constant.
\end{enumerate}

\noindent \textbf{3.\ Application:} First, we systematically assess the practical expressivity of various 1-Lip architectures based on function fitting, Wasserstein-1 distance estimation and Wasserstein GAN training.
Then, as our main application, we perform image reconstruction within the popular PnP framework.
Here, we also prove that using 1-Lip networks leads to the stability of the data-to-reconstruction map.
Our framework significantly outperforms the others for PnP image reconstruction, and at least matches them in all other experiments.
Hence, we expect that LLS can be successfully deployed to any learning task that requires 1-Lip architectures.
Our code is accessible on Github\footnote{\href{https://github.com/StanislasDucotterd/Lipschitz_DSNN}{https://github.com/StanislasDucotterd/Lipschitz{\_}DSNN}}.

\section{1-Lip Neural Networks}

A function $f \colon \mathbb{R}^{m} \rightarrow \mathbb{R}^{n}$ is $K$-Lipschitz ($K > 0$) if, for all $\mathbf{x}_1, \mathbf{x}_2$ $\in \mathbb{R}^{m}$, it holds that
\begin{equation}
\|f(\mathbf{x}_1)-f(\mathbf{x}_2)\| \leq K  \|\mathbf{x}_1-\mathbf{x}_2\|.
\end{equation}
The Lipschitz constant $\operatorname{Lip}(f)$ of $f$ is the smallest constant $K$ such that $f$ is $K$-Lipschitz. Here, we only consider $\|\cdot\|$ to be the 2-norm (also known as the Euclidean norm).
Complementary to our framework, there also exists a line of work focusing on the $\infty$-norm setting instead \citep{madry_towards_2019, zhang_towards_2021, zhang_rethinking_2022}.

In this paper, we consider feedforward neural networks $f_{\theta}\colon \mathbb{R}^{N_0} \rightarrow \mathbb{R}^{N_L}$ of the form
\begin{equation}\label{eq:architecture}
f_{\theta}(\mathbf{x}) = A_{L} \circ \cdots \circ \sigma_{\ell} \circ A_{\ell} \circ \cdots \circ \sigma_{1} \circ A_{1}(\mathbf{x}),
\end{equation}
where each $A_{\ell}\colon \mathbb{R}^{N_{\ell-1}} \rightarrow \mathbb{R}^{N_{\ell}}$, $\ell = 1,\ldots,L$, is a linear layer given by
\begin{equation}
A_{\ell}(\mathbf{x}) = \mathbf{W}_{\ell}\mathbf{x} + \mathbf{b}_{\ell},
\end{equation}
with weight matrices $\mathbf{W}_{\ell} \in \mathbb{R}^{N_{\ell}, N_{\ell-1}}$ and bias vectors $\mathbf{b}_{\ell} \in \mathbb{R}^{N_{\ell}}$.
The model incorporates fixed or learnable nonlinear activation functions
$\sigma_{\ell}\colon \mathbb{R}^{N_{\ell}} \rightarrow \mathbb{R}^{N_{\ell}}$. For component-wise activation functions, we have that $\sigma_\ell(\vec x) = (\sigma_{\ell, n}(x_n))_{n=1}^{N_\ell}$ with individual scalar activation functions $\sigma_{\ell, n}\colon \mathbb{R} \rightarrow \mathbb{R}$. The complete set of parameters of the network is denoted by $\theta$.

A straightforward way to enfore $\operatorname{Lip}(f_{\theta}) \leq 1$ is to use the sub-multiplicativity of the Lipschitz constant for the composition operation, which yields the estimate

\begin{equation}
\operatorname{Lip}(f_{\theta}) \leq \operatorname{Lip}(A_L) \prod_{\ell=1}^{L-1} \operatorname{Lip}(\sigma_{\ell}) \operatorname{Lip}(A_{\ell}).
\end{equation}
Consequently, it suffices to constrain the Lipschitz constant of each $A_{\ell}$ and $\sigma_{\ell}$ by 1. 

\subsection{1-Lipschitz Linear Layers}\label{lip_linear_layers}

It is known that the Lipschitz constant of the linear layer $A_{\ell}$ is equal to the largest singular value of its weight matrix $\mathbf{W}_{\ell}$.
In our experiments, we constrain $\mathbf{W}_{\ell}$ in two ways.

\begin{itemize}
    \item {\bf Spectral Normalization:} This method rescales each linear layer $A_{\ell}$ by dividing its weight matrix $\mathbf{W}_{\ell}$ by its largest singular value. The latter is estimated via power iterations.
    This method was introduced for fully connected networks in \cite{MKKY2018} and later generalized for convolutional layers in \cite{ryu2019plug}.
    \item {\bf Orthonormalization:} 
    Here, the $\vec W_\ell$ are forced to be orthonormal, so that $\vec W_\ell^T \vec W_\ell$ is the identity matrix. Unlike spectral normalization, which only constrains the largest singular value, this method forces all the singular values to be one.
    Various implementations of orthonormalization have been proposed to handle both fully connected \citep{ALG19} and convolutional layers \citep{li_preventing_2019, su_scaling-up_2022}.
\end{itemize} 

\subsection{1-Lipschitz Activation Functions}\label{lipschitz_activation_section}

Here, we shortly introduce all $1$-Lipschitz activation functions that we compare against LLS.

\begin{itemize}
\item {\bf ReLU:} The activation function $\operatorname{ReLU}(\vec x) = (\max(0, x_n))_{n=1}^N$ acts component-wise.
\item {\bf Absolute Value:} The absolute value (AV) activation function is component-wise and GNP. It is given by $\operatorname{AV}(\vec x) = (|x_n|)_{n=1}^N$.
\item {\bf Parametric ReLU:} The parametric ReLU (PReLU) activation function \citep{he_delving_2015} acts component-wise.
It is given by $\operatorname{PReLU}_{\boldsymbol{a}}(\mathbf{\vec x}) = (\max(a_n x_n, x_n))_{n=1}^{N}$ with learnable parameters $(a_n)_{n=1}^N$.
Since $\Lip(\operatorname{PReLU}_{\boldsymbol{a}}) = \max(\max_{1\leq n\leq N} |a_n|, 1)$, an easy way to make it 1-Lipschitz is to clip the parameters $(a_n)_{n=1}^N$ in $[-1 ,1]$.
\item {\bf GroupSort:} This activation function \citep{ALG19} separates the pre-activations into groups of size $k$ and sorts each group in ascending order.
Hence, it is locally a permutation and therefore GNP.
If the group size is 2, GroupSort (GS) is called MaxMin.
1-Lip MaxMin and GS neural networks are universal approximators for $1$-Lipschitz functions in a specific setting where the first weight matrix satisfies $\|\mathbf{W}_1\|_{2, \infty} \leq 1$ and all other weight matrices satisfy $\|\mathbf{W}_l\|_{\infty}\leq1$ \citep[Theorem 3]{ALG19}. 
\item {\bf Householder:} The householder (HH) activation function \citep{SSF2021} separates the pre-activations into groups of size 2, and for any $\vec x \in \mathbb{R}^2$, computes

\begin{equation}
\operatorname{HH}_{\mathbf{v}}\left(\mathbf{x}\right)= \begin{cases}
\mathbf{x}, & \mathbf{v}^T\mathbf{x} > 0 \\
\left(\mathbf{I} - 2\mathbf{vv}^T\right)\mathbf{x}, & \mathbf{v}^T\mathbf{x} \leq 0, \end{cases}
\end{equation}
where $\vec v \in \mathbb{R}^2$ with $\|\mathbf{v}\| = 1$ is learnable. The HH activation function is GNP.
\end{itemize}
For these choices, Proposition~\ref{prop:SameExp} holds. The proof is given in Appendix~\ref{sec:SameExp}.
\begin{proposition}\label{prop:SameExp}
    On any compact set $D \subset \R^{N_0}$, 1-Lip neural networks with AV, PReLU, GS, or HH activation functions can represent the same set of functions.
\end{proposition}

By contrast to Proposition~\ref{prop:SameExp}, 1-Lip ReLU networks are less expressive and can only represent a subset of these functions.

\section{1-Lip Learnable Linear Spline Networks}

It has been shown in \cite{Unser2019} that neural networks with LLS activation functions are the solution of a functional optimization problem that consists of the optimization of a neural network with freeform activation functions under a second-order total-variation constraint.
\cite{BCGA2020} proposed a way to learn the linear splines and to efficiently control their effective number of linear regions via a regularization term in the training loss.
They propose a fast implementation with a computational complexity that does not depend on the number of linear regions of the LLS.
As a first step toward Lipschitz-constrained LLS networks, \cite{AGCU2020} added a term in the training loss that penalizes a loose bound of the Lipschitz constant of the LLS activation functions.
This approach, however, does not offer a strict control of the overall Lipschitz constant of the network.

In this section, we extend the reasoning and implementation to the strict 1-Lip setting. 
Ideally, we want to train a neural network with 1-Lipschitz linear layers and freeform \mbox{1-Lipschitz} activation functions.
Unfortunately, this leads to a difficult infinite-dimensional optimization problem.
In order to promote simple solutions, we use the second-order total variation as regularizer, which favors activation functions with sparse second-order derivatives while ensuring differentiablity almost everywhere. 

\subsection{Representer Theorem and Expressivity}

The second-order total variation of a function $f$ is defined as
\begin{equation}
\TV{2}(f) = \|\mathrm{D}^2f\|_{\mathcal{M}},
\end{equation}
where $\mathrm{D}$ is the distributional derivative operator and $\|\cdot\|_{\mathcal{M}}$ is the total-variation norm. For any function $f$ in the space $L_1(\mathbb{R})$ of absolutely integrable functions, it holds that $\|f\|_{\mathcal{M}} = \|f\|_{L_1}$. However, unlike the $L_1$ norm, the total-variation norm is also well-defined for any shifted Dirac impulse with $\|\delta(\cdot-\tau)\|_{\mathcal{M}} = 1$, $\tau \in \mathbb{R}$ (see Appendix~\ref{sec:tv2} for technical details).
In the sequel, we consider activation functions in the space $\operatorname{BV}^{(2)}(\mathbb{R}) = \{f: \mathbb{R} \rightarrow \mathbb{R} \text{ s.t. } \TV{2}(f) < +\infty\}$ of functions with bounded second-order total variation.

Given a series $(\mathbf{x}_m, \mathbf{y}_m)$, $m = 1,\ldots,M$, of data points and a neural network $f_{\theta}\colon \mathbb{R}^{N_0} \to \mathbb{R}^{N_{L}}$ with $f_{\theta} = \sigma_{L} \circ g_\theta$, where $g_\theta \colon \mathbb{R}^{N_0} \to \mathbb{R}^{N_{L}}$ has architecture \eqref{eq:architecture}, we propose the constrained regularized training problem
\begin{equation}\label{eq:optim_network}
\begin{aligned}
&\argmin_{\substack{\mathbf{W}_{\ell}, \mathbf{b}_{\ell}, \sigma_{\ell, n} \in \operatorname{BV}^{(2)}(\mathbb{R})\\\text{s.t.} \operatorname{Lip}(\sigma_{\ell, n}) \leq 1, \, \|\mathbf{W}_{\ell}\| \leq 1}} \left(\sum_{m=1}^{M} E\bigl(\mathbf{y}_{m},  f_{\theta}\left(\mathbf{x}_{m}\right)\bigr)+\lambda \sum_{\ell=1}^{L} \sum_{n=1}^{N_{\ell}} \TV{2}\left(\sigma_{\ell, n}\right)\right),
\end{aligned}
\end{equation}
where $E\colon \mathbb{R}^{N_{\ell}} \times \mathbb{R}^{N_{\ell}} \to \mathbb{R}^{+}$ is proper, lower-semicontinuous, and
coercive. Theorem~\ref{thm:spline_good1} states that a neural network with linear-spline activation functions suffices to find a solution of \eqref{eq:optim_network}.
Its proof can be found in Appendix~\ref{sec:spline_good1}.

\begin{theorem}\label{thm:spline_good1}
A solution of \eqref{eq:optim_network} always exists and can be chosen as a neural network with activation functions of the form
\begin{equation}\label{eq:LearnSpline}
    \sigma_{\ell, n}(x) = b_{1, \ell, n}+b_{2, \ell, n} x+\sum_{k=1}^{K_{\ell, n}} a_{k, \ell, n}\operatorname{ReLU}\left(x-\tau_{k, \ell, n}\right)
\end{equation}
with $K_{\ell, n} \leq M-2$, knots $\tau_{1, \ell, n}, \ldots, \tau_{K_{\ell, n}, \ell, n} \in \mathbb{R}$,  scalar biases  $b_{1, \ell, n}, b_{2, \ell, n} \in \mathbb{R}$, and weights $a_{1, \ell, n}, \ldots, a_{K_{\ell, n}, \ell, n} \in \mathbb{R}$.
\end{theorem}

This result, which is similar to the representer theorems in \cite{Unser2019} and \cite{AGCU2020}, shows that there exists an optimal solution with linear-spline activation functions.
The important point in the statement of the theorem is that each neuron has its own free parameters, including the (a priori unknown) number of knots $K_{\ell, n}$, the determination of which is part of the training procedure. Beside the strict control of the Lipschitz constant, which is not covered by the representer theorems in \citep{Unser2019}, a noteworthy improvement brought by Theorem~\ref{thm:spline_good1} is the existence of a solution, which is still an open problem in the unconstrained case.
To this end, we have assumed that the last layer of the neural network $f_{\theta}$ consists of an activation function $\sigma_{L}$ only.
This theoretical setup for the proof is not a strong restriction since the freeform activation $\sigma_L$ has the possibility to be the identity mapping in practice. 

The second-order total variation of the activation functions in \eqref{eq:LearnSpline} is given by
\begin{align}
\label{eq:control_knots}
    \|\mathrm{D}^2 \sigma_{\ell, n}\|_{\mathcal{M}} &= \biggl\|\sum_{k=1}^{K_{\ell, n}} a_{k, \ell, n}\delta\left(\cdot-\tau_{k, \ell, n}\right)\biggr\|_{\mathcal{M}} = \sum_{k=1}^{K_{\ell, n}} |a_{k, \ell, n}|\|\delta(\cdot - \tau_{k, \ell, n})\|_{\mathcal{M}}\notag\\ &=\sum_{k=1}^{K_{\ell, n}} |a_{k, \ell, n}| = \|\mathbf{a}_{\ell, n}\|_1,
\end{align}
where we used that $\mathrm{D}^2\operatorname{ReLU}(\cdot - \tau) = \delta(\cdot - \tau)$ and $\mathrm{D}^2\{b_1 + b_2 x\} = 0$ for all $\tau, b_1, b_2 \in \mathbb{R}$.
The idea of LSS networks is to have learnable activation functions of the form $\eqref{eq:LearnSpline}$.

There is an approximation result for 1-Lip neural networks with LLS activation functions \citep[Theorem 4.3]{neumayer2022approximation} that states that, when the LLSs have three linear regions, they already achieve the optimal expressivity among all $1$-Lip neural networks with component-wise activation functions. 
The proof relies on the fact that the number of knots can be decreased by the addition of layers. For this reason, it is unclear whether it is always sufficient in practice to have only three linear regions for a given architecture. 
Remarkably, our framework allows us to parameterize each LLS activation function $\sigma_{\ell, n}$ with more linear regions and to then sparsify them during the training process through $\TV{2}$ regularization.
Indeed, equation~\eqref{eq:control_knots} shows that the latter amounts to a penalization of the term $\|\vec a_{\ell, n}\|_1$, which will favor solutions with fewer linear regions. 

%Notice that the absolute value is a linear spline, therefore, it follows from Proposition \ref{prop:SameExp} that any 1-Lip neural network with any of the activation functions defined in Section \ref{lipschitz_activation_section} can be represented exactly by a 1-Lip DSNN. It is unknown whether the converse is true, as each of those activation functions only has two linear regions.

% \begin{theorem}[\cite{neumayer2022approximation}]
% \label{thm:spline_good2}
%  Let $D \subset \mathbb{R}^{N_0}$ be compact.
%  Then neural networks $f\colon D \rightarrow \mathbb{R}^{N_L}$ with architecture \eqref{eq:architecture}, $2$-norm constrained weights, and 1-Lipschitz linear spline activation functions with 3 linear regions can approximate the same functions as the corresponding neural networks $g\colon D \rightarrow \mathbb{R}^{N_L}$ with arbitrary 1-Lipschitz component-wise activation functions.
% \end{theorem}

\subsection{Deep Spline Neural Network Representation}
\label{sec:deepspline}

The parameterization \eqref{eq:LearnSpline} has two drawbacks when training neural networks.

\begin{itemize}
    \item The learning of the number and positions of the knots is challenging.
    \item The evaluation time is linear in the number of ReLUs.
\end{itemize}
Our implementation differs from the two main frameworks \citep{agostinelli2014learning, jin_deep_2016} by evading those two drawbacks through the use of localized basis functions $\varphi_{T}(x)=\beta^{1}(x/T)$ on a uniform grid with stepsize $T$ and $k_{\text{max}} - k_{\text{min}} + 1$ knots, as proposed by \cite{BCGA2020}, where $\beta^1$ is the B-spline of degree one defined as
\begin{equation}
\beta^{1}(x) = \begin{cases}1-|x|, & x \in[-1,1] \\
0, & \text {otherwise.}\end{cases}
\end{equation}
To ease the presentation, we describe the implementation of LLS networks in Sections~\ref{sec:deepspline} and \ref{sec:methods} for a single LLS activation function $\sigma$ expressed as
\begin{equation}
\sigma(x)= \begin{cases}c_{k_{\min }}+\frac{1}{T}\left(c_{k_{\min }}-c_{k_{\min }-1}\right)\left(x-k_{\min } T\right), & x \in\left(-\infty, k_{\min } T\right) \\ \sum_{k=k_{\min}-1}^{k_{\max }} c_{k} \beta^1(x/T-k), & x \in\left[k_{\min } T, k_{\max } T\right] \\ c_{k_{\max }}+\frac{1}{T}\left(c_{k_{\max }+1}-c_{k_{\max }}\right)\left(x-k_{\max } T\right), & x \in\left(k_{\max } T, \infty\right).\end{cases}
\end{equation}
For any $x \in \mathbb{R}$, the computation of $\sigma(x)$ requires the evaluation of at most two basis functions. The activation function $\sigma$ is nonlinear on $\left[k_{\min } T, k_{\max } T\right]$ and extrapolated linearly outside of this interval. It is fully described by the stepsize $T$ and by a vector $\mathbf{c} \in \mathbb{R}^K$ with $K = k_{\text{max}} - k_{\text{min}} + 3$. It has $\operatorname{Lip}(\sigma) = \frac{1}{T}\|\mat D \vec c\|_{\infty}$, where $\mathbf{D} \in \mathbb{R}^{K-1,K}$ is the first-order finite-difference matrix.

\begin{figure}[t]
    \centering
    \includegraphics[scale=0.9]{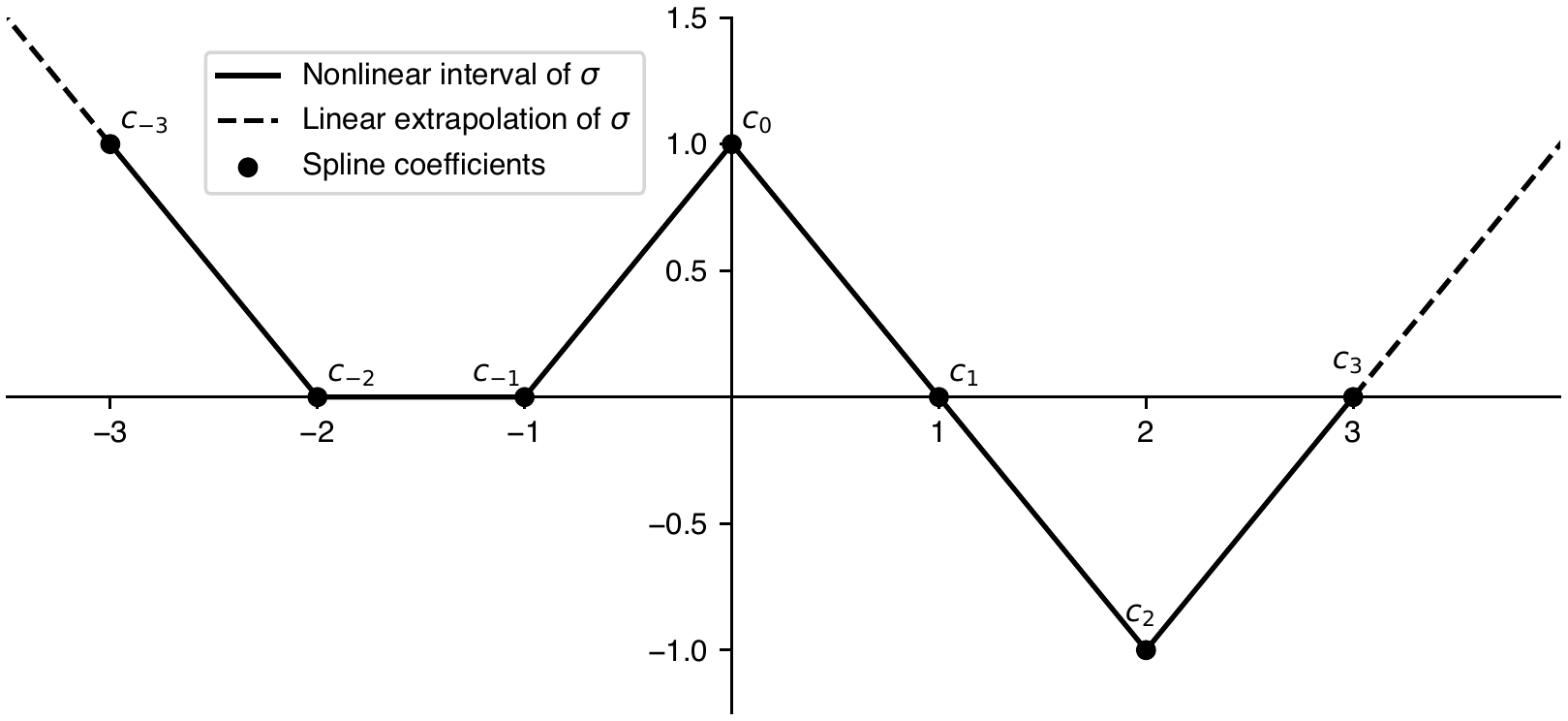}
    \caption{A LLS activation function with $T=1$ and $K=7$. The function is nonlinear in $[-3, 3]$, and linearly extrapolated outside. At $x=1$, the second-order finite-difference is zero, which effectively removes one linear region.
    This behavior is favored by the regularization term $\frac{1}{T}\Vert \mathbf{Lc}\Vert_1$ which promotes sparse second-order finite differences and thereby decreases the number of effective linear regions.}
    \label{fig:spline_activation}
\end{figure}

In practice, we choose a high number $K$ and a small stepsize $T$.
We then ensure that a simple activation function is learned by using $\TV{2}$ regularization. In our setting, $\TV{2}(\sigma) = \Vert \mathbf{a}\Vert_1 = \frac{1}{T}\Vert\mathbf{Lc}\Vert_1$, where $\mathbf{L}$ is the second-order finite-difference matrix. Overall, we impose strict bounds on the first-order finite differences of the coefficients $\vec c$, and we seek to sparsify their second-order finite differences. This procedure leads to an approximate learning of the optimal position of each knot for each 1-Lipschitz LLS. An illustration of a possible $\sigma$ is shown in Figure \ref{fig:spline_activation}.

Within LLS networks, the LLS can be initialized in many ways, some including popular activation functions such as ReLU, leaky ReLU, PReLU, or MaxMin.
Further, the LLS activation functions can also be shared, which saves memory and training cost, and allows one to have one activation function per channel in convolutional neural networks.

\subsection{Methods}
\label{sec:methods}

To ensure that every activation function $\sigma$ is 1-Lipschitz, the absolute difference between any two consecutive coefficients must be at most $T$.
Hence, the corresponding set of feasible coefficients is given by $\{\mathbf{c} \in \mathbb{R}^K: \|\mathbf{Dc}\|_{\infty} \leq T\}$. A first attempt at a minimization over this set has been made in \cite{bohra2021learning}.
There, the authors use a method that divides each activation function by its maximum slope after each training step.
In Section \ref{sec:SplineProj}, we present an alternative projection scheme that is better suited to optimization and yields a much better performance in practice, while being just as fast. Additionally, we introduce a scaling parameter for each activation function, which facilitates the training and increases the performance of the network even further at a negligible computational cost.

\subsubsection{Constrained Coefficients}\label{sec:SplineProj}

The textbook approach to maintain the 1-Lipschitz property throughout an iterative minimization scheme would be to determine the least-squares projection onto $\{\mathbf{c} \in \mathbb{R}^K: \|\mathbf{Dc}\|_{\infty} \leq T\}$ at each iteration.
This operation would preserve the mean of $\vec c$, as shown in Appendix \ref{sec:splineproj_app}.
Unfortunately, its computation is very expensive as it requires to solve a quadratic program after each training step and for each activation function.
As substitute, we introduce a simpler projection SplineProj that also preserves  the mean while being much faster to compute. 
In brief, SplineProj computes the finite-differences, clips them, sums them and adds a constant to the preservation of the mean.

Let us denote the Moore--Penrose pseudoinverse of $\mat D$ by $\mat D^\dagger$ and the vector of ones by $\vec 1 \in \mathbb{R}^K$.
Further, we require the component-wise operation
\begin{equation}\label{eq:slope_clipping}
\operatorname{Clip}_T(x) = \begin{cases} -T, & x < -T \\
x, & x \in [-T, T] \\
T, & x > T.\end{cases}
\end{equation}

\begin{proposition}\label{prop:splineproj}
    The operation $\operatorname{SplineProj}$ defined as
    \begin{equation}\label{slope_clipping}
    \operatorname{SplineProj}(\mathbf{c}) = \mathbf{D}^\dagger \operatorname{Clip}_T(\mathbf{Dc}) + \mathbf{1}\frac{1}{K}\sum_{k=1}^K c_k
\end{equation}
has the following properties:
\begin{enumerate}
    \item it is a projection onto the set $\{\vec c \in \mathbb{R}^K: \|\mat D \vec c\|_\infty \leq T\}$;
    \item it is almost-everywhere differentiable with respect to $\vec c$;
    \item it preserves the mean of $\vec c$.
\end{enumerate} 
\end{proposition}

The proof of Proposition \ref{prop:splineproj} can be found in Appendix \ref{sec:splineproj_app}.

%Practically, SplineProj translates into a very efficient implementation in Pytorch using the \texttt{mean} and \texttt{cumsum} methods.
In gradient-based optimization, one usually handles domain constraints by projecting the variables back onto the feasible set after each gradient step.
However, this turned out to be inefficient for neural networks in our experiments.
Instead, we parameterize the LLS activation functions directly with $\operatorname{SplineProj}(\mathbf{c})$, which leads to unconstrained training.
This strategy is in line with the popular spectral normalization of \cite{MKKY2018}, where the weight matrices are unconstrained and parameterized using an approximate projection.
For our parameterization approach, Property 2 of Proposition \ref{prop:splineproj} is very important as it allows us to back-propagate through SplineProj during the optimization process.
To compute SplineProj efficiently, we calculate $\mat D^\dagger$ in a matrix-free fashion with a cumulative sum. The computational cost of SplineProj is negligeable compared to the cost of constraining the linear layer to be 1-Lipschitz.

\subsubsection{Scaling Parameter}\label{sec:ScalePara}
%\textbf{Old version:} A potential limitation of our B-spline parameterization is that the activation functions can only be nonlinear on the range $[T k_{\text{min}}, T k_{\text{max}}]$ which is set a priori.
%In neural networks, different neurons may play different roles and can have different input distributions. We propose to adapt our parameterization by inserting an additional trainable scaling factor. Specifically, we create a new activation function $\tilde{\sigma}$ out of $\sigma$ as \\
We propose to increase the flexibility of our LLS activation functions by the introduction of an additional trainable scaling factor $\alpha$. Specifically, we propose the new activation function
\begin{equation}
    \tilde \sigma (x) = \frac{1}{\alpha} \sigma(\alpha x).
\end{equation}
With this scaling, $\tilde \sigma$ is nonlinear on $[k_{\text{min}} T/\alpha, k_{\text{max}} T/\alpha]$ and the Lipschitz constant
\begin{equation}
    \operatorname{Lip}(\tilde \sigma) = \sup_{x_1, x_2 \in \mathbb{R}} \frac{|\frac{1}{\alpha}\sigma(\alpha x_1) - \frac{1}{\alpha}\sigma(\alpha x_2)|}{|x_1 - x_2|} = \sup_{x_1, x_2 \in \mathbb{R}} \frac{\frac{1}{\alpha}|\sigma(\alpha x_1) - \sigma(\alpha x_2)|}{\frac{1}{\alpha}|\alpha x_1 - \alpha x_2|} = \operatorname{Lip}(\sigma)
\end{equation}
is left unchanged.
As detailed in Appendix \ref{sec:tv2}, the second-order total variation is preserved as well.
Basically, $\alpha$ allows us to decrease the data-fitting term defined in \eqref{eq:optim_network} without breaking the constraints or increasing the complexity of the activation functions.
Experimentally, we indeed found that the performance of LSS networks improves if we also optimize over $\alpha$.
In contrast, the ReLU, AV, PReLU, GS, and HH activation functions are invariant to this parameter and do not benefit from it.
In practice, the scaling parameter $\alpha$ is initialized as one and updated via standard stochastic gradient-based methods. Throughout our experiments, every LLS activation function has its own scaling parameter $\alpha$.

\section{Experiments}

With our experiments in Section~\ref{sec:EvalExp}, we gauge the expressivity of 1-Lip architectures that use either LLS or one of the activation functions from Section~\ref{lipschitz_activation_section}.
Moreover, we assess the relevance of our proposed learning strategy for LLS.
After this validation study, we benchmark our $1$-Lip architectures for the practical use-case of PnP image reconstruction in Section~\ref{sec:PnPReconstruction}.
There, we also prove important stability guarantees.

All 1-Lip networks are learned with the Adam optimizer \citep{KingmaB14} and the default hyperparameters of its PyTorch implementation.
For the parameters of the PReLU and HH activation functions, the learning rate is the same as for the weights of the network.
The LLS networks use three different learning rates: $\eta$ for the weights, $\eta$/4 for the scaling parameters $\alpha$, and $\eta$/40 for the remaining parameters of the LLS. 
These ratios remain fixed throughout this section and, hence, only $\eta$ is going to be stated.

\subsection{Evaluating the Expressivity}\label{sec:EvalExp}
\subsubsection{One-Dimensional Function Fitting}

\begin{figure}[t]
    \centering
    \includegraphics[scale=0.9]{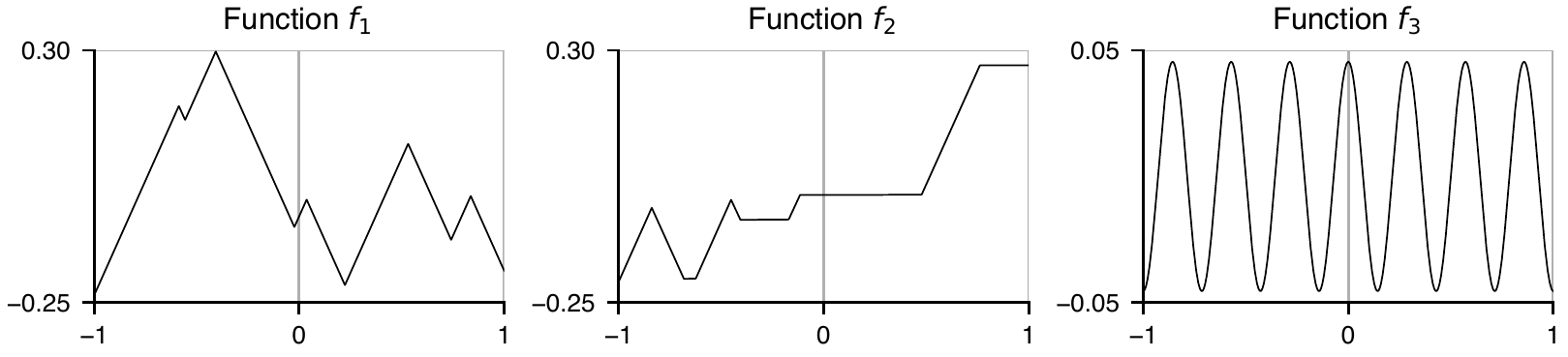}
    \caption{Three 1-Lipschitz functions that we attempt to fit with 1-Lip neural networks.
    All functions have zero mean over the interval $[-1, 1]$.}
    \label{fig:1d_functions}
\end{figure}
Here, we use 1-Lip networks to fit the three 1-Lipschitz functions $f_i\colon [-1,1] \to \mathbb{R}$ in Figure~\ref{fig:1d_functions}.
With this experiment, we aim to assess if the architectures achieve the full expressivity in the class of 1-Lipschitz functions on the real line.
For $f_1$, we have $\vert \nabla f_1 \vert = 1$ almost everywhere.
Hence, the GNP activation functions are expected to perform well and serve as a baseline against which we compare LLSs.
The function $f_2$ alternates between $\vert \nabla f_2 \vert = 1$ and $\vert \nabla f_2 \vert = 0$.
It was designed to test the ability of LLS networks to fit functions with constant regions.
Lastly, we benchmark all methods on the highly varying function $f_3(x) = \sin(7\pi x)/7\pi$, which is challenging to fit under Lipschitz constraints.
Additionally, we probe the impact of the two methods described in Sections~\ref{sec:SplineProj} and \ref{sec:ScalePara} on the performance of the LLS networks by comparing the proposed implementation (denoted as LLS New) with the one from \cite{bohra2021learning} (denoted as LLS Old), which relies on simple normalization.

Each network comes in two variants, namely with orthonormalization and spectral normalization of the weights.
The mean squared error (MSE) loss is computed over 1000 uniformly sampled points from $[-1,1]$ for training, and a uniform partition of $[-1, 1]$ with 10000 points for testing.
For each instance, we tuned the width, the depth, and the hyperparameters of the network for the smallest test loss.
ReLU networks have 10 layers and a width of 50; AV, PReLU, and HH networks have 8 layers and a width of 20; GS networks have 7 layers and a width of 20; LSS networks have 4 layers and a width of 10.
We initialized the PReLU as the absolute value, we used GS with a group size of 5, and the LLS was initialized as ReLU and had a range of $[-0.5, 0.5]$, 100 linear regions, and  $\lambda=10^{-7}$ for the $\TV{2}$ regularization.
Every network relied on Kaiming initialization \citep{he_delving_2015} and was trained 25 times with a batch size of 10 for 1000 epochs. 
The LLS networks always used $\eta = \num{2e-3}$, while the other ones used $\eta = \num{4e-3}$ for $f_1$, $f_2$ and $\eta = 10^{-3}$ for $f_3$.

\begin{figure}[tb]
    \centering
    \includegraphics[scale=0.85]{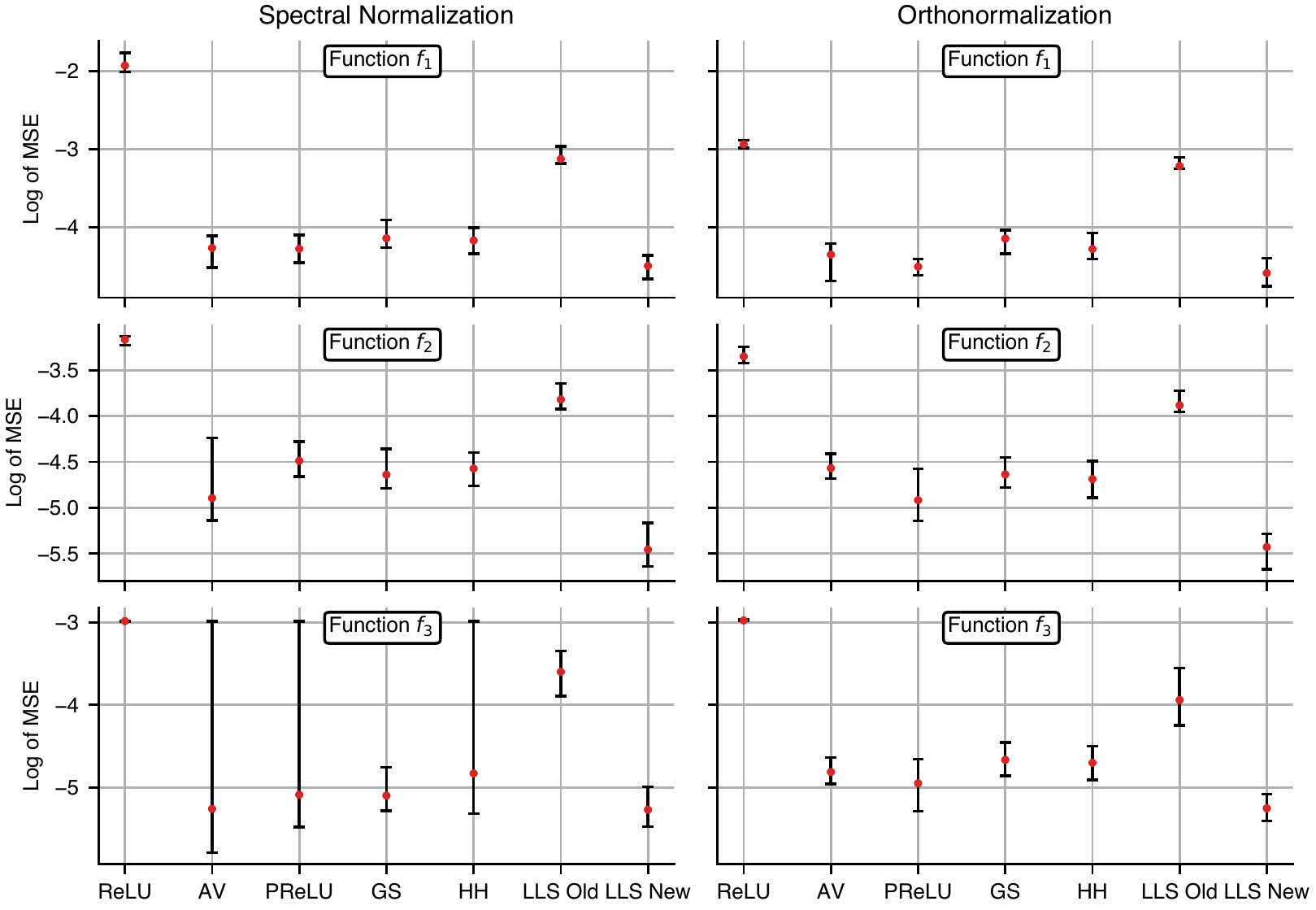}
    \caption{Fitting results for the functions from Figure \ref{fig:1d_functions}.
    The red markers represent the median performance. The black bars represent the lower and upper quartiles, respectively.}
    \label{fig:1d_performances}
\end{figure}

We report the median and the two quartiles of the test losses in Figure~\ref{fig:1d_performances}.
For the spectral normalization, it appears that AV, PReLU, and HH tend to get stuck in local minima when fitting $f_3$ (the associated upper quartile of the loss is quite large).
In return, we observe that LLS consistently outperforms the other architectures in all experiments.
Particularly striking is the improvement of LLS New over LLS Old, which confirms the benefits of the two modules described in Sections \ref{sec:SplineProj} and \ref{sec:ScalePara}.
Accordingly, from now on, we drop LLS Old and only retain LLS New. 

\subsubsection{High-Dimensional Function Fitting: Wasserstein Distances}

The Wasserstein-1 distance $W_1$ is a metric for probability distributions.
It has been used by \cite{pmlr-v70-arjovsky17a} to improve the performance of GANs, which were first introduced in \cite{goodfellow_generative_2014}.
Using the Kantorovich dual formulation \citep{villani_optimal_2016}, we can compute $W_1$ by solving an optimization problem over the space of 1-Lipschitz functions
\begin{equation}\label{wasserstein}
    W_1(P_1, P_2) = \sup _{\operatorname{Lip}(f) \leq 1} \mathbb{E}_{\vec x \sim P_{1}}[f(\vec x)]-\mathbb{E}_{\vec x \sim P_{2}}[f(\vec x)].
\end{equation}
In \eqref{wasserstein}, we can use a neural representations to parameterize the $f$ for optimization purposes.
Since expressive architectures are very important in this context, we get another good benchmark.
Experimentally, \cite{ALG19} observed that orthonormalization of the linear layers is superior to spectral normalization for this task.
Hence, we only use the former in our experiments.
Further, \cite{GAAD2017} have shown that, under reasonable assumptions, any $f^*$ that maximizes \eqref{wasserstein} satisfies $\vert \nabla f^*\vert = 1$ almost everywhere.
Hence, GNP architectures are expected to perform better.
In general, estimating $W_1$ for high-dimensional distributions is very challenging, see \cite{KorKolBur2022} for a recent survey.

\paragraph{Gaussian Mixtures}

In our first setting, the distributions are

\begin{equation}
    P_1 = X_1 Z_{1} + (1 - X_1) Z_{2} \quad \text{ and } \quad P_2 = X_2 Z_{3} + (1 - X_2) Z_{4},
\end{equation}
where $\mathbb{P}(X_k = 0) = \mathbb{P}(X_k = 1) = 1/2$ for $k = 1,2$ and $Z_k \sim \mathcal{N}(\mathbf{\mu}_k, \mathbf{\Sigma}_k)$ for $k = 1,...,4$.
The $\mathbf{\mu}_k \in \R^N$ and $\mathbf{\Sigma}_k \in \R^{N,N}$, $N \in\{5, 10, 20\}$, are random with $(\mathbf{\mu}_k)_n \sim \mathcal{N}(0, 1)$ and $\mathbf{\Sigma}_k = \mathbf{A}_k^T\mathbf{A}_k$ with $(\mathbf{A}_k)_{nm} \sim \mathcal{N}(0, 1)$ and $1\leq n,m\leq N$.
For each instance, we tuned the width and the depth of the fully connected neural representation for best performance.
Irrespective of $N$, ReLU architectures had 30 layers and a width of 1024; all other architectures had 10 layers and a width of 2048.
The PReLUs were initialized as the ReLU for $N \in \{5,10\}$, and as MaxMin for $N = 20$.
We used GS with a group size of 2 for $N \in \{ 5, 20\}$ and 4 for $N = 4$.
The LLS had 50 linear regions for $N = 5$ and 100 for $N \in \{10, 20\}$.
Their range was $[-1, 1]$, $[-5, 5]$, and $[-10, 10]$ for $N =$ 5, 10 and 20, respectively.
Further, we set $\lambda=10^{-10}$ for the $\TV{2}$ regularization, and initialized the LLS as the ReLU for $N \in \{5, 10\}$ and as MaxMin for $N = 20$.
The additional LLS coefficients increased the architecture parameters by at most 0.7$\%$.
All neural representations used orthogonal initialization \citep{DBLP:journals/corr/SaxeMG13} and were optimized for 10000 gradient steps with $\eta = \num{5e-3}$ and batches of 4096 samples.

In Table~\ref{tab:wasserstein_bigaussian}, we report the mean and standard deviation over five runs of the Monte Carlo estimation for $W_1$ as in \eqref{wasserstein} with the learned $f$ and $10^5$ samples.
ReLU leads to an estimate that is significantly lower than the others, which is, most likely, due to its lack of expressivity.
Otherwise, the estimates are quite similar.
LLS has the best estimate for $N \in\{5,10\}$ but is slightly outperformed for $N = 20$.

\begin{table}[tb]
\centering
\caption{Estimated Wasserstein distance for the Gaussian mixtures.}%
\label{tab:wasserstein_bigaussian}
\small
\setlength\tabcolsep{4.5pt}
\begin{tabular}{l|cccccc}
\toprule
$N$ & ReLU & AV & PReLU & GS & HH & LLS \\%& Ground Truth\\
\midrule
5 & 2.009/0.004 & 2.271/0.006 & 2.279/0.006 & 2.271/0.006 & 2.272/0.006 & \textbf{2.283}/0.006 \\%& 2.443\\
10 & 5.960/0.014 & 6.461/0.011 & 6.465/0.012 & 6.475/0.012 & 6.461/0.012 & \textbf{6.486}/0.011\\%& 7.665\\
20 & 11.638/0.007 & 13.187/0.012 & 13.245/0.012 & \textbf{13.251}/0.012 & 13.247/0.012 & 13.243/0.012 \\%& 18.311\\
\bottomrule
\end{tabular}
\end{table}

\paragraph{MNIST}

Here, $P_1$ is a uniform distribution over a set of real MNIST\footnote{\href{http://yann.lecun.com/exdb/mnist/}{http://yann.lecun.com/exdb/mnist/}} images and $P_2$ is the generator distribution of a GAN trained to generate MNIST images.
The architecture of this GAN is taken from \cite{chen_infogan_2016}. 
All neural representations of $f$ are fully connected with a width of 1024, and various depths.
They were optimized 5 times each for 2000 epochs with $\eta = \num{2e-3}$ and orthogonal initialization \citep{DBLP:journals/corr/SaxeMG13}. 
For a depth of 3, GS has group size of 8, and PReLU and LLS were initialized as the absolute value.
For a depth of 5 or 7, GS has a group size of 2, and PReLU and LLS were initialized as MaxMin. 
The LLS have a range of $[-0.15, 0.15]$, 20 linear regions, and $\lambda=10^{-10}$.
Their coefficients only increase the total number of parameters in the architecture by 2\%.
We optimize the neural representation on 54000 images from the MNIST training set and use the 6000 remaining ones as validation set.
The test set contains 10000 MNIST images.

In Table~\ref{tab:wasserstein_results}, we report the estimated $W_1$ metric between the MNIST images of the test set and the ones generated by the GAN.
%For this large-scale problem it is practically unfeasible to compute the true $W_1$ metric between the samples.
Again, ReLU leads to an estimate that is significantly lower than the others.
Otherwise, the results are more or less similar except for AV and HH with depth 7 and 3, respectively, which are worse than the others. 

\begin{table}[tb]
\centering
\caption{Mean and standard deviation of the estimated Wasserstein distance over five trials for several architectures.}%
\label{tab:wasserstein_results}
\small
\setlength\tabcolsep{4.5pt}
\begin{tabular}{l|ccccccc}
\toprule
Depth & ReLU & AV & PReLU & GS & HH & LLS \\
\midrule
3 & 0.727/0.001 & \textbf{1.190/0.002} & \textbf{1.190/0.002} & $1.189/0.001$ & 1.165/0.001 & \textbf{1.190/0.002}\\
5 & $0.881/0.001$ & $1.368/0.003$ & $1.371/0.002$ & $1.369/0.002$ & $1.369/0.002$ & \textbf{1.373/0.003}\\
7 & 0.960/0.001 & $1.406/0.008$ & $1.437/0.002$ & $1.436/0.001$ & \textbf{1.440/0.003} & $1.439/0.001$\\
\bottomrule
\end{tabular}
\end{table}

\subsubsection{1-Lipschitz Wasserstein GAN Training}\label{sec:WGAN}
We train Wasserstein GANs to generate MNIST images.
To this end, we use the same framework and optimization process as \cite{ALG19}, where the discriminators have strict Lipschitz constraints instead of the commonly used relaxation in terms of a gradient penalty.
On the contrary, the generators themselves are unconstrained.
Thus, we use ReLUs as their activation functions and only plug the activation functions from Section~\ref{lipschitz_activation_section} into the discriminator.
Here, GS has a group size of 2, PReLU was initialized as MaxMin, LLS was initialized as the absolute value, has range $[-0.1, 0.1]$, 20 linear regions, and $\lambda=10^{-6}.$ The spline coefficients only increase the total number of parameters in the neural network by 0.2\%. The Wasserstein GANs were trained on the MNIST training set.

\begin{figure}
    \centering
    \includegraphics[scale=0.825]{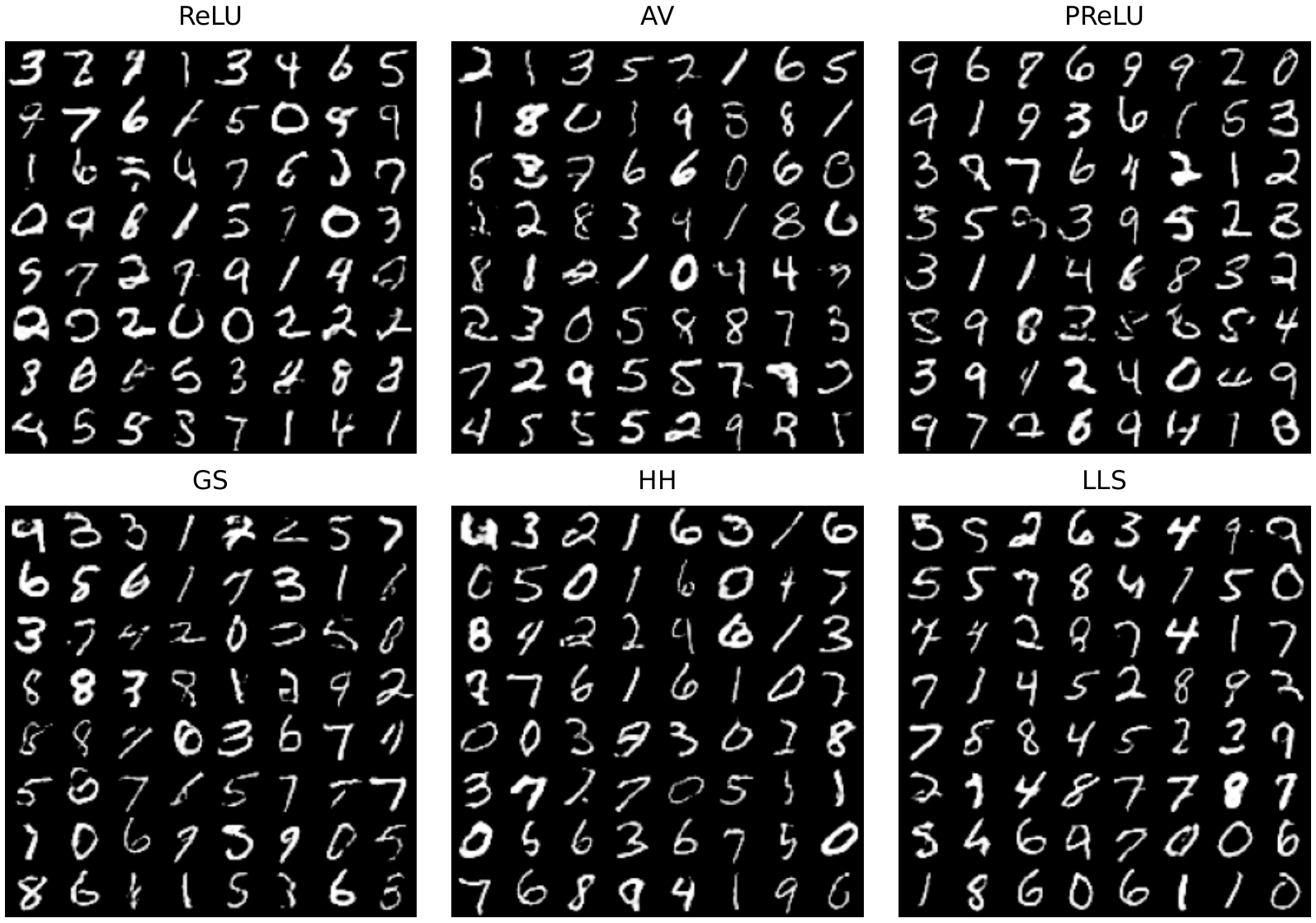}
    \caption{Digits generated by the Wasserstein GANs with different activation functions.}
    \label{fig:gan_samples}
\end{figure}

We report the inception score on the MNIST test set using the implementation from \cite{li_alice_2017} in Table~\ref{tab:inception_scores}.
As expected, the limited expressivity of the ReLU leads to the lowest score.
LLS yields the best score of all schemes and its ability to generate realistic digits can be appreciated visually in Figure~\ref{fig:gan_samples}.
Still, we should keep in mind that the main purpose of this experiment is an expressivity comparison and not generative performance.

\begin{table}[tb]
\centering
\caption{Inception scores for MNIST digit generation.}%
\label{tab:inception_scores}
\small
\begin{tabular}{l|ccccccc}
\toprule
 & ReLU & AV & PReLU & GS & HH & LLS \\
\midrule
Inception score & 1.88 & 2.19 & 2.13 & 2.17 & 2.07 & \textbf{2.38}\\
\bottomrule
\end{tabular}
\end{table}

\subsection{Image Reconstruction via Plug-and-Play}\label{sec:PnPReconstruction}

Many image-reconstruction tasks can be formulated as a linear inverse problem.
Specifically, the task is to recover an image $\mathbf{x} \in \mathbb{R}^n$ from the noisy measurement
\begin{equation}
\label{first_equ}
    \mathbf{y} = \mathbf{Hx} + \mathbf{n} \in \mathbb{R}^m,
\end{equation}
where $\mathbf{H} \in \mathbb{R}^{m \times n}$ is a measurement operator and $\mathbf{n} \in \mathbb{R}^{m}$ is random noise.
Problem \eqref{first_equ} is nondeterministic and often ill-posed, in the sense that multiple images yield the same measurements.
To make \eqref{first_equ} well-posed, one usually incorporates prior knowledge about the unknown image $\vec x$ by adding regularization.
This leads to the reconstruction problem
\begin{equation}
\label{eq:obj_function}
    \min_{\mathbf{x} \in \mathbb{R}^{n}} f(\mathbf{y}, \mathbf{H}\mathbf{x}) + g(\mathbf{x}),
\end{equation}
where $f\colon \R^m \times \R^m \to \R^+$ is a data-fidelity term and $g\colon \R^n \to \R^+$ is a prior that favors certain types of solutions.
If $f$ is differentiable and $g$ is convex, \eqref{eq:obj_function} can be minimized by the iterative forward-backward splitting (FBS) algorithm \citep{CW05} with

\begin{equation}
    \mathbf{x}^{k+1}=\prox_{\alpha g}\bigl(\mathbf{x}^k-\alpha \pmb{\nabla} f(\mathbf{y}, \mathbf{Hx}^{k})\bigr).
\end{equation}
Here, the proximal operator of $g$ is defined as $\prox_{g}(\mathbf{z})= \argmin_{\mathbf{x} \in \R^{d}} \tfrac12 \|\mathbf{x}-\mathbf{z}\|^{2} + g(\mathbf{x})$.
The idea behind PnP algorithms \citep{VBW13} is to replace $\prox_{\alpha g}$ with a generic denoiser $D\colon \mathbb{R}^n \to \mathbb{R}^n$.
While not necessarily corresponding to an explicit regularizer $g$, this approach has led to improved results compared to conventional methods as it allows the use of powerful deep-learning-based denoisers, as done in \cite{ryu2019plug, sun_scalable_2021, ye_deep_2018}.
The convergence of the PnP-FBS iterations
\begin{equation}
\label{eq:pnp_fbs}
    \mathbf{x}^{k+1}=D\bigl(\mathbf{x}^k-\alpha \pmb{\nabla} f(\mathbf{y}, \mathbf{Hx}^{k})\bigr)
\end{equation}
can be guaranteed \cite[Proposition 15]{hertrich_convolutional_2021} if 
\begin{itemize}
    \item $D$ is averaged, i.e., of the form $D = \beta R + (1 - \beta)\operatorname{Id}$ with a 1-Lipschitz $R$ and $\beta \in(0, 1)$;
    \item $f(\mathbf{y}, \mathbf{H}\cdot)$ is convex, differentiable with $L$-Lipschitz gradient, and $\alpha \in (0, 2/L)$.
\end{itemize}
In addition, we now prove the Lipschitz continuity of the data-to-reconstruction map, which is an important property for image reconstruction methods.
\begin{proposition}\label{prop:stability1}
Let $\vec x_1^*$ and $\vec x_2^*$ be fixed points of \eqref{eq:pnp_fbs} for measurements $\vec y_1$ and $\vec y_2$, respectively.
If $D$ is averaged with $\beta \leq 1/2$ and $f(\mathbf{y}, \mathbf{Hx}) = \frac{1}{2}\|\mathbf{y} - \mathbf{Hx}\|_2^2$, then it holds that
\begin{equation}\label{eq:EstForward}
\|\vec H\vec x_1^* - \vec H\vec x_2^*\| \leq \|\vec y_1 - \vec y_2\|.
\end{equation}
\end{proposition}
If $\vec H$ is invertible, this yields the direct relation 
\begin{equation}\label{eq:H_invert}
\|\vec x_1^* - \vec x_2^*\| \leq \frac{1}{\sigma_{\text{min}}(\vec H^T \vec H)}\|\vec y_1 - \vec y_2\|.
\end{equation}
Under slightly stronger constraints on $D$, we also have a result for non-invertible $\mat H$.
\begin{proposition}\label{prop:stability2}
    In the setting of Proposition \ref{prop:stability1}, it holds for $K$-Lipschitz $D$, $K < 1$, that
    \begin{equation}\label{eq:EstCont}
    \|\vec x_1^* - \vec x_2^*\| \leq        \frac{\alpha \|\vec H\|K}{1 - K}\|\vec     y_1 - \vec y_2\|.
    \end{equation}
\end{proposition}

Propositions \ref{prop:stability1} and \ref{prop:stability2} are proven in Appendix~\ref{sec:pnp_results}.
In principle, the model \eqref{eq:obj_function} leads to better data consistency than the one provided by the end-to-end neural network frameworks that directly reconstruct $\mathbf{x}$ from $\mathbf{y}$.
Those latter approaches are also known to suffer from stability issues \citep{antun_instabilities_2020} and, more importantly, have been found to remove or hallucinate structure \citep{nataraj, muckley_results_2021}, which is unacceptable in diagnostic imaging.
In sharp contrast, our PnP approach \eqref{eq:pnp_fbs} comes with the stability estimates \eqref{eq:EstForward}, \eqref{eq:H_invert} and \eqref{eq:EstCont} which typically do not hold for other PnP methods.
%Moreover, it ensures data consistency if $\beta \leq 1/2$.

\subsubsection{Learning a Denoiser for PnP}
A good denoising network is the backbone of most PnP methods.
Unfortunately, most common architectures \citep{ronneberger_u-net_2015,zhang_beyond_2017,liang_swinir_2021} are not natively 1-Lipschitz.
They rely on dedicated modules designed to improve the denoising performance, such as skip connections, batch normalization, and attention modules.
These make it challenging to build provably averaged denoisers.
Instead, we use a plain CNN architecture that is equivalent to the DnCNN of \cite{zhang_beyond_2017} without the residual.
This architecture is easy to constrain and still provides competitive performance.
In detail, we train 1-Lip denoisers that are composed of 8 orthogonal convolutional layers and the activation functions from Section \ref{lipschitz_activation_section}.
The convolutional layers are parameterized with the BCOP framework \citep{li_preventing_2019} and have kernels of size $(3 \times 3)$.
For the LLS, we take 64 channels.
To compensate for the additional spline parameters, we train every other model with 68 channels.
 
The training dataset consists of 238400 patches of size $(40 \times 40)$ taken from the BSD500 image dataset \citep{arbelaez_contour_2011}.
All images take values in $[0, 1]$.
We train denoisers for Gaussian noise with standard deviation $\sigma =  5/255, 10/255, 15/255$.
All denoisers are trained for 50 epochs with a batch size of 128, $\eta = \num{4e-5}$, and the MSE loss function.
The PReLU activation functions were initialized as the absolute value. GS has a group size of 2. The LLS activation functions have 50 linear regions, a range of 0.1, and were initialized as the identity.
In this experiment, we also investigated the effect of the $\TV{2}$ regularization parameter $\lambda$ on the performance and the number of linear regions in the LLSs.
The performances on the BSD68 test set are provided in Table~\ref{table:performance}.
For each noise level, LLS performs best, and, as expected, ReLU is doing worse than all the others.

The number of linear regions for the LLS $\sigma_{\ell, n}$ is equal to $\frac{1}{T}\|\mathbf{Lc}_{\ell, n}\|_0 + 1$.
This metric can overestimate the number of linear regions due to numerical imprecisions. Instead, we define the effective number of linear regions as $(|\{1 \leq k \leq K_{\ell, n}: |(\mat L \vec c_{\ell, n})_k| > 0.01\}| + 1)$. For each LSS network, we report in Table~\ref{table:nbr_knots} the average number of effective linear regions (AELR) over all the $\sigma_{\ell, n}$.
An AELR close to one indicates that the majority of neurons become skip connection, which corresponds to a simplification of the network. Without regularization, the $\sigma_{\ell, n}$ have an AELR of 8.07 to 9.24 out of the 50 available linear regions. The $\TV{2}$ regularization drastically sparsifies the $\sigma_{\ell, n}$.
With $\lambda \in[10^{-6}, 10^{-4}]$, the AELR is between 1.07 and 1.44, which is a large decrease without degradation in the denoising performances. 
For $\lambda=10^{-3}$, the $\sigma_{\ell, n}$ are even further sparsified at the cost of a small loss of performance. 
We observe a significant loss of performance when $\lambda$ is increased to $10^{-2}$, where the network is almost an affine mapping.
Notice that the AELR is 2 for ReLU and AV, meaning that LLS outperforms them while being simpler.
Another interesting observation is that, despite being very sparse on average, the LLS networks with $\lambda \in[10^{-6}, 10^{-3}]$ have at least one $\sigma_{\ell, n}$ with at least three linear regions.
This suggests that most of the common activation functions might be suboptimal as they have only two linear regions.

\begin{table}[tb]
\centering
\caption{PSNR and SSIM values on BSD68 for each activation function and noise level.}
\label{table:performance}
\begin{tabular}{lcccccc}
\toprule
Noise level & \multicolumn{2}{c}{ $\sigma = 5/255$ } & \multicolumn{2}{c}{ $\sigma = 10/255$ } & \multicolumn{2}{c}{ $\sigma = 15/255$ } \\
Metric & PSNR & SSIM & PSNR & SSIM & PSNR & SSIM \\
\midrule
ReLU & 36.10 & 0.9386 & 31.92 & 0.8735 & 29.76 & 0.8203 \\
AV & 36.58 & 0.9499 & 32.33 & 0.8889 & 30.09 & 0.8375 \\
PReLU & 36.58 & 0.9498 & 32.25 & 0.8887 & 30.11 & 0.8367\\
GS & 36.54 & 0.9489 & 32.23 & 0.8845 & 30.11 & 0.8346 \\
HH & 36.47 & 0.9476 & 32.25 & 0.8866 & 30.11 & 0.8350 \\
LLS ($\lambda = 0$) & 36.85 & 0.9540 & \textbf{32.59} & \textbf{0.8978} & 30.35 & 0.8464 \\
LLS ($\lambda = 10^{-6}$) & \textbf{36.86} & \textbf{0.9546} & 32.55 & 0.8962 & \textbf{30.38} & \textbf{0.8479} \\
LLS ($\lambda = 10^{-5}$) & \textbf{36.86} & 0.9543 & 32.55 & 0.8960 & 30.34 & 0.8455 \\
LLS ($\lambda = 10^{-4}$) & 36.82 & 0.9534 & 32.57 & 0.8970 & 30.36 & 0.8468 \\
LLS ($\lambda = 10^{-3}$) & 36.63 & 0.9497 & 32.47 & 0.8924 & 30.31 & 0.8437 \\
LLS ($\lambda = 10^{-2}$) & 35.15 & 0.9142 & 32.00 & 0.8782 & 29.73 & 0.8156 \\
\bottomrule
\end{tabular}
\vspace{.2cm}

\caption{Average number of effective linear regions (AELR) for several $\lambda$ and noise levels.
}
\label{table:nbr_knots}
\vspace{.1cm}
\begin{tabular}{lcccccc}
\toprule
Noise level & $\lambda = 0$ & $\lambda = 10^{-6}$ & $\lambda = 10^{-5}$ & $\lambda = 10^{-4}$ & $\lambda = 10^{-3}$ & $\lambda = 10^{-2}$\\
\midrule
$\sigma = 5/255$ & 9.24 & 1.21 & 1.11 & 1.07 & 1.02 & 1.00 \\
$\sigma = 10/255$ & 8.76 & 1.24 & 1.15 & 1.14 & 1.06 & 1.01\\
$\sigma = 15/255$ & 8.07 & 1.44 & 1.24 & 1.25 & 1.10 & 1.02\\
\bottomrule
\end{tabular}
\end{table}

\subsubsection{Numerical Results for PnP-FBS}\label{sec:exp_pnp}
Now, we want to deploy the learned denoisers $D_{\sigma}$ (which are learned with $\lambda = 10^{-6}$ for LLS) in the PnP-FBS algorithm for image reconstruction, where we use the data-fidelity term $f(\mathbf{y}, \mathbf{Hx}) = \frac{1}{2}\|\mathbf{y} - \mathbf{Hx}\|_2^2$.
To ensure convergence, we set $\alpha = 1/ \|\mat H^* \mat H\|$, and we tune the noise level $\sigma$ of $D_{\sigma}$ on the validation set over $\sigma = 5/255, 10/255, 15/255$. 
To further adapt the denoising strength and to make them $\beta$-averaged, we replace the $D_{\sigma}$ by $D_{\sigma,\beta}=\beta D_{\sigma} + (1-\beta)\operatorname{Id}$, where the parameter $\beta\in[0,1)$ is also tuned on the validation set.
In our experiments, we actually noticed that the best $\beta$ is always lower than 1/2, which means that the conditions in Proposition \ref{prop:stability1} are met.
As the GS activation function involves sorting of the feature map along its 68 channels, it takes significantly more time than the other activation functions.
Hence, it is impractical to tune the hyperparameters, and we do not use it for this image reconstruction experiment.

\paragraph{Single-Coil MRI}
Here, we want to recover $\vec x$ from $\vec y=\mathbf{M F x}+\mathbf{n} \in \mathbb{C}^{M}$, where $\mathbf{M}$ is a subsampling mask (identity matrix with some missing entries), $\mathbf{F}$ is the discrete Fourier transform matrix, and $\mathbf{n}$ is a realization of a complex-valued Gaussian noise characterized by $\sigma_\mathbf{n}$ for the real and imaginary parts.
This noise level is not to be confused with the noise level $\sigma$ that appears in $D_{\sigma,\beta}$.
We use fully sampled knee MR images of size $(320 \times 320)$ from the fastMRI dataset \citep{zbontar2018fastMRI} as ground truths.
Specifically, we create validation and test sets consisting of 100 and 99 images, respectively, which are individually normalized within the range $[0,1]$.
For our experiments, the subsampling mask $\mathbf{M}$ is specified by two parameters: the acceleration $M_{\text{acc}}$ and the center fraction $M_{\text{cf}}$.
It selects the $\lfloor 320 M_{\text{cf}} \rfloor$ columns in the center of the k-space (low frequencies).
Further, it selects columns uniformly at random from the other regions in the k-space such that the total number of selected columns is $\lfloor 320/M_{\text{acc}} \rfloor$.
The measurements are simulated with $\sigma_\mathbf{n} = 0.01$. 

\begin{figure}[tpb]
    \centering
    \includegraphics[scale=0.835]{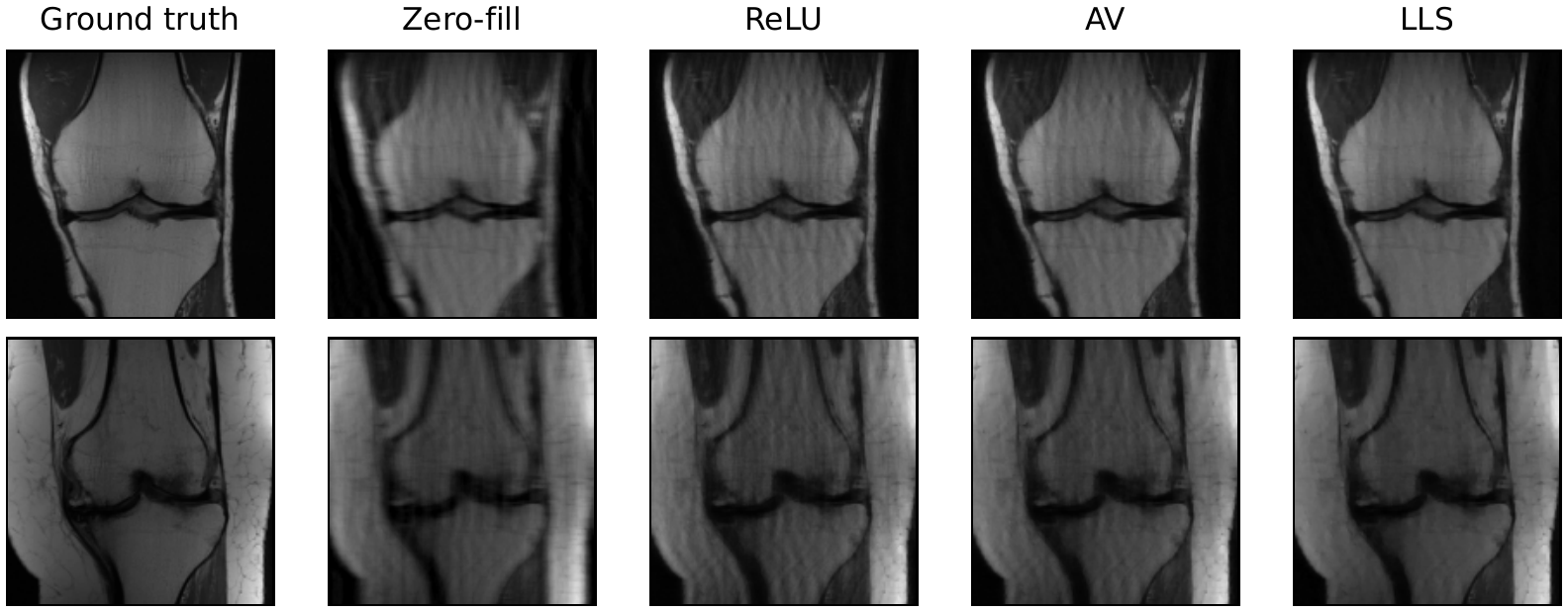}
    \caption{The ground truth, the zero-fill reconstruction $\vec H^T \vec y$, and the PnP-FBS reconstruction using networks with ReLU, LLS, and AV for the single-coil MRI experiment.
    %AV has the second best performance after LLS in terms of PSNR.
    }
    \label{fig:mri}
    \vspace{.35cm}
    
    \centering
    \includegraphics[scale=0.42]{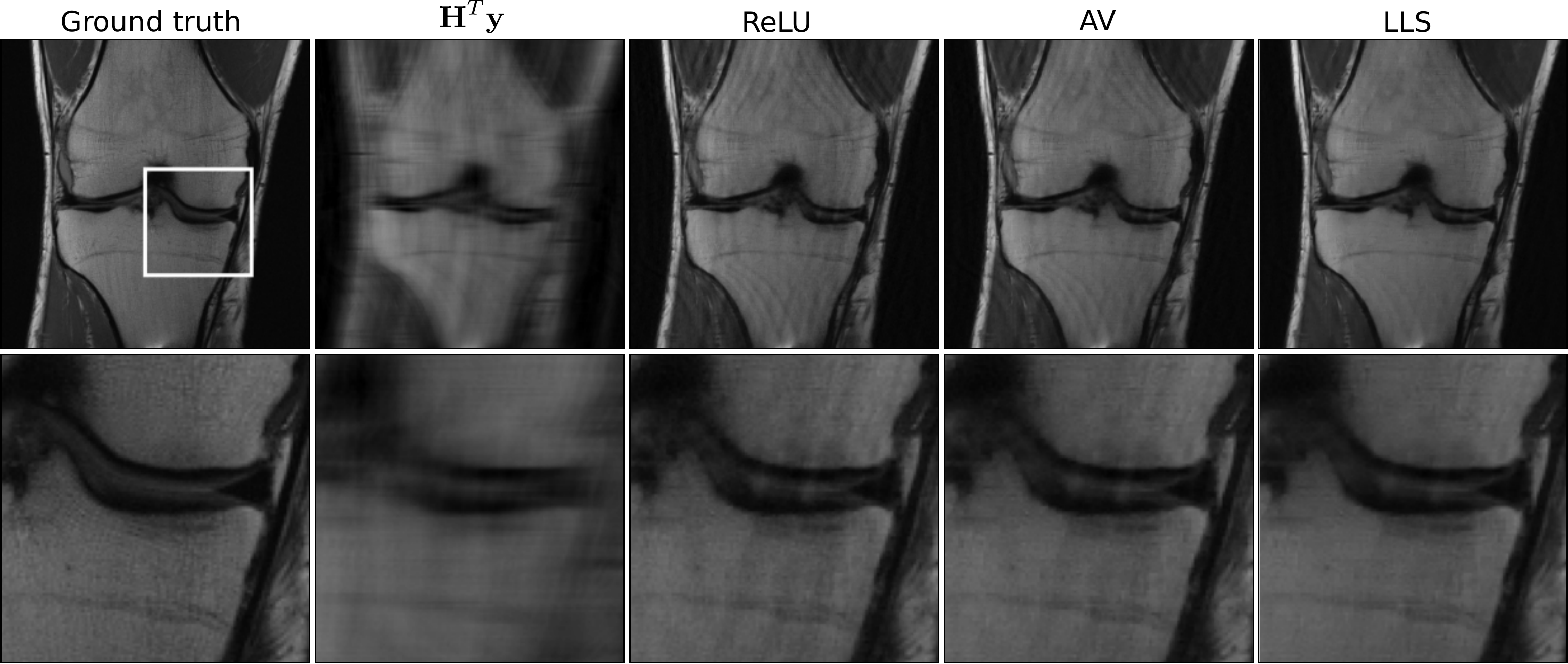}
    \caption{The ground truth, the zero-fill reconstruction $\vec H^T \vec y$, and the PnP-FBS reconstruction using networks with ReLU, LLS, and AV for the multi-coil MRI experiment.
    %AV has the second best performance after LLS in terms of PSNR.
    }
    \label{fig:multi_mri}
    \vspace{.35cm}
    
    \centering
    \includegraphics[scale=0.42]{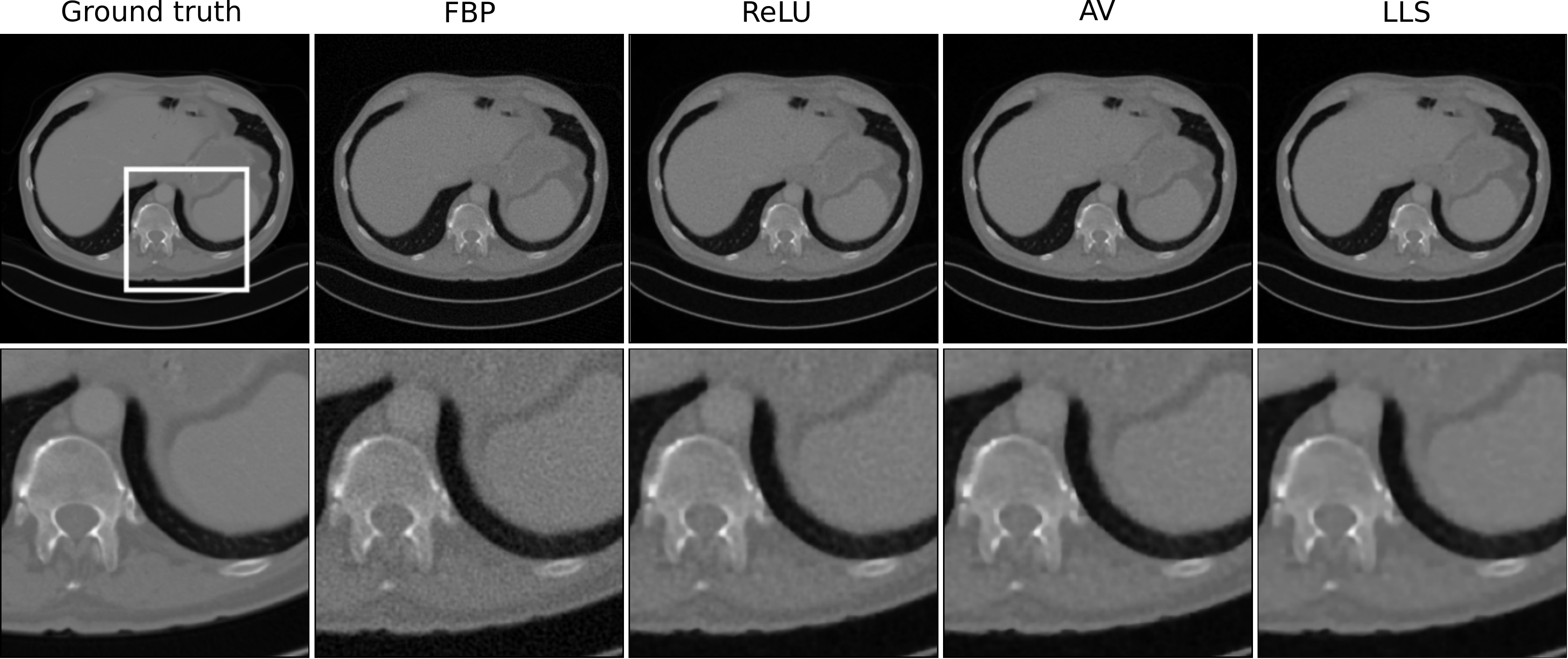}
    \caption{Reconstructions for the CT experiment. We report the ground truth, the filtered backprojection  and the PnP-FBS reconstruction using networks with ReLU, LLS, and AV.
    %AV has the second best performance after LLS in terms of PSNR.
    }
    \label{fig:ct}
\end{figure}

The reconstruction performances over the test set are reported in Table~\ref{table:reconstruction_performance}.
We observe a significant gap between LLS and the other activation functions for both masks $M$ in terms of PSNR and SSIM.
Actually, LLS outperforms the other schemes on every single image from the test set.
In Figure~\ref{fig:mri}, we observe stripe-like structures in the zero-fill reconstruction.
These are typical aliasing artifacts that result from the subsampling in the horizontal direction in Fourier space.
They are significantly reduced in the LLS reconstruction.

\paragraph{Multi-Coil MRI with 15 Coils}

Here, the data is given by $\vec y = (\vec y_1,\ldots, \vec y_{15})$ with $\vec{y}_k = \vec {MFS}_k\vec x+\mathbf{n}_k$ and complex-valued diagonal matrices $\vec S_k \in \mathbb{C}^{n \times n}$ (called sensitivity maps).
These maps were estimated from the data $\vec y$ using the ESPIRiT algorithm \citep{Uecker2014-uv}, which is part of the BART toolbox \citep{uecker2013software}.
Again, the ground-truth images are from the fastMRI dataset \citep{zbontar2018fastMRI}, but both with (PDFS) and without (PD) fat suppression.
The Fourier subsampling is performed with the parametric Cartesian mask from before.
Gaussian noise with standard deviation $\sigma_{\mathbf{n}} = \num{2e-3}$ is added to the real and imaginary parts of the measurements.
The reconstruction performances over the test set are reported in Table~\ref{table:reconstruction_performance_mri}.
Again, we observe a significant gap in terms of PSNR and SSIM between LLS and the other activation functions for both masks.
LLS outperforms the other schemes on every single image from the test set.

\paragraph{Computed Tomography (CT)}

The groundtruth comes from human abdominal CT scans for 10 patients provided by Mayo Clinic for the low-dose CT Grand Challenge \citep{M2016}.
The validation set consists of 6 images taken uniformly from the first patient of the training set from \cite{mukherjee_learned_2021}.
We use the same test set as \cite{mukherjee_learned_2021}, more precisely, 128 slices with size $(512\times 512)$ that correspond to one patient.
The data $\vec y$ is simulated through a parallel-beam acquisition geometry with 200 angles and 400 detectors.
These measurements are corrupted by Gaussian noise with standard deviation $\sigma_{\mathbf{n}}\in\{ 0.5, 1, 2\}$. 
The reconstruction performance in terms of PSNR and SSIM over the test set are reported in Table~\ref{table:reconstruction_performance_ct}.
Again, we observe a significant gap between LLS and the other activation functions.
LLS outperforms the other schemes on every single image from the testing set.
Reconstructions for one image are reported in Figure~\ref{fig:ct}.

\begin{table}[tp]
\centering
\caption{PSNR and SSIM values for the single-coil MRI reconstruction experiment.}%
\label{table:reconstruction_performance}
\begin{tabular}{lcccc}
\toprule
($M_{\text{acc}}$, $M_{\text{cf}}$) & \multicolumn{2}{c}{(4, 0.08)} & \multicolumn{2}{c}{(6, 0.06)} \\
Metric & PSNR & SSIM & PSNR & SSIM \\
\midrule
Zero-fill & 27.55 & 0.6895 & 25.55 & 0.6223\\
ReLU & 29.97 & 0.7574 & 26.87 & 0.6781\\
AV & 30.61 & 0.7721 & 27.35 & 0.6921\\
PReLU & 30.58 & 0.7716 & 27.32 & 0.6906\\
HH & 30.55 & 0.7696 & 27.34 & 0.6887 \\
%LLS ($\lambda = 0$)& 31.41  & 0.7899 & 27.96  & 0.7091 \\
LLS  & \textbf{31.54} & \textbf{0.7924} & \textbf{28.04}  & \textbf{0.7108} \\
%LLS ($\lambda = 10^{-5}$) & 31.52  & 0.7919  & 28.05  & \textbf{0.7111}  \\
%LLS ($\lambda = 10^{-4}$) & \textbf{31.55}  & \textbf{0.7924}  & \textbf{28.06}  & 0.7109  \\
\bottomrule
\end{tabular}
\vspace{.3cm}

\centering
\caption{PSNR and SSIM values for the multi-coil MRI experiment.}%

\label{table:reconstruction_performance_mri}
\setlength\tabcolsep{4.5pt}
\begin{tabular}{lcccccccc}
\toprule
($M_{\text{acc}}$, $M_{\text{cf}}$) & & \multicolumn{2}{c}{(4, 0.08)} & & & \multicolumn{2}{c}{(8, 0.04)} \\ & \multicolumn{2}{c}{PSNR} & \multicolumn{2}{c}{SSIM} & \multicolumn{2}{c}{PSNR} & \multicolumn{2}{c}{SSIM} \\ & PD & PDFS & PD & PDFS & PD & PDFS & PD & PDFS \\
\midrule
Zero-fill & 27.71 & 29.94 & 0.751 & 0.759 & 23.80 & 27.19 & 0.648 & 0.681\\
ReLU & 37.21 & 37.06 & 0.929 & 0.915 & 31.37 & 32.57 & 0.837 & 0.822 \\
AV & 37.81 & 37.48 & 0.935 & 0.919 & 31.82 & 32.95 & 0.845 & 0.829 \\
PReLU & 37.71 & 37.51 & 0.934 & 0.919 & 31.67 & 33.11 & 0.845 & 0.832 \\
HH & 37.66 & 37.39 & 0.933 & 0.919 & 31.68 & 32.91 & 0.843 & 0.829 \\
LLS & \textbf{38.68} & \textbf{37.96} & \textbf{0.943} & \textbf{0.924} & \textbf{32.75} & \textbf{33.61} & \textbf{0.859} & \textbf{0.835} \\
\bottomrule
\end{tabular}
\vspace{.3cm}

\centering
\caption{PSNR and SSIM values for the CT experiment.}%

\label{table:reconstruction_performance_ct}
\setlength\tabcolsep{4.5pt}
\begin{tabular}{lcccccc}
\toprule
& \multicolumn{2}{c}{$\sigma_{\vec n}$=0.5} & \multicolumn{2}{c}{$\sigma_{\vec n}$=1} & \multicolumn{2}{c}{$\sigma_{\vec n}$=2} \\
& PSNR & SSIM & PSNR & SSIM & PSNR & SSIM  \\
\midrule
FBP & 32.14 & 0.697 & 27.05 & 0.432 & 21.29 & 0.204 \\
ReLU & 36.94 & 0.914 & 33.65 & 0.860 & 30.34 & 0.782 \\
AV & 37.15 & 0.926 & 34.19 & 0.885 & 31.07 & 0.813 \\
PReLU & 37.18  & 0.927 & 34.21 & 0.887 & 30.87 & 0.812 \\
HH & 36.94 & 0.918 & 34.11 & 0.877 & 30.92 & 0.809 \\
LLS & \bf{38.19} & \bf{0.931} & \bf{35.15} & \bf{0.897} & \bf{31.85} & \bf{0.844} \\
\bottomrule
\end{tabular}
\end{table}
\section{Conclusion}
In this paper, we proposed a framework to efficiently train $1$-Lipschitz neural networks with learnable linear-spline activation functions.
First, we formulated the training stage as an optimization task and showed that the solution set contains networks with linear-spline activation functions.
Our implementation of this framework embeds the required $1$-Lipschitz constraint on the splines directly into the forward pass.
Further, we added learnable scaling factors, which preserve the Lipschitz constant of the splines and enhance the overall expressivity of the network.
For the practically relevant PnP image reconstruction, our approach significantly outperforms $1$-Lipschitz architectures that rely on other (popular) activation functions such as the parametric ReLU and Householder.
In this setting, classical choices such as ReLU suffer from limited expressivity and should not be used.
Our observations are a starting point for the exploration of other architectural constraints and learnable non-component-wise activation functions within the framework of $1$-Lipschitz networks.

\section*{Acknowledgment}
The research leading to these results was supported by the European Research Council (ERC) under European Union’s Horizon 2020 (H2020), Grant Agreement - Project No 101020573 FunLearn and by the Swiss National Science Foundation, Grant 200020 184646/1.

\bibliography{references}
\appendix

\section{Proof of Proposition \ref{prop:SameExp}}\label{sec:SameExp}

\label{proof:SameExp} We express the activation functions in terms of each other using weights $\vec W_k$ with $\Vert \vec W_k \Vert_2 \leq 1$.
Choose $B$ such that $x + B > 0$ for all $x \in D$ and any pre-activation in the network.

\paragraph{AV as Expressive as PReLU:} We can express AV using PReLU with $a = -1$.
For the other direction, we have that
\begin{align}
&\operatorname{PReLU}_{a}(x)\notag \\
=&\begin{bmatrix}
\sqrt{(1+a)/2} & -\sqrt{(1-a)/2}
\end{bmatrix}
\operatorname{AV}\left(\begin{bmatrix}
\sqrt{(1+a)/2} \\
\sqrt{(1-a)/2}
\end{bmatrix} x + \begin{bmatrix}
\sqrt{(1+a)/2}B \\
0
\end{bmatrix}\right) - \frac{1+a}{2B}.
\end{align}

\paragraph{AV as Expressive as GS:} This was already proven in \cite{ALG19}, but we include the expressions for the sake of completeness.
It holds that
\begin{equation}
\begin{bmatrix}
\max (x_1) \\
\min (x_2)
\end{bmatrix}=\mathbf{M} \operatorname{AV}\left(\mathbf{M}\begin{bmatrix}
x_1 \\
x_2
\end{bmatrix}+\begin{bmatrix}
B \\
0
\end{bmatrix}\right)-\begin{bmatrix}
\sqrt{2} B \\
0
\end{bmatrix},
\end{equation}
where
\begin{equation}
    \mathbf{M}=\frac{1}{\sqrt{2}}\begin{bmatrix}
1 & 1 \\
1 & -1
\end{bmatrix}.
\end{equation}
For the reverse direction, we have that
\begin{equation}
\operatorname{AV}(x)=\begin{bmatrix}
\frac{1}{\sqrt{2}} & -\frac{1}{\sqrt{2}}
\end{bmatrix} \operatorname{MaxMin}\left(\begin{bmatrix}
\frac{1}{\sqrt{2}} \\
\frac{1}{\sqrt{2}}
\end{bmatrix} x\right).
\end{equation}

\paragraph{GS as Expressive as HH:}
For $\mathbf{v} = \frac{1}{\sqrt{2}}(1, -1)$ we have that $\text{HH}_\mathbf{v} = \text{MaxMin}$.
Further, we can also express $\text{HH}_\mathbf{v}$ using MaxMin as

\begin{equation}
    \operatorname{HH}_{\mathbf{v}}\left(\mathbf{z}\right) = \mathbf{R}(\mathbf{v})\operatorname{MaxMin}\bigl(\mathbf{R}(\mathbf{v})^T\mathbf{z}\bigr),
\end{equation}
where $\mathbf{R}(\mathbf{v})$ is the rotation matrix 
\begin{equation}
\mathbf{R}(\mathbf{v})=\begin{bmatrix}
\cos \gamma(v_1, v_2) & -\sin \gamma(v_1, v_2) \\
\sin \gamma(v_1, v_2) & \cos \gamma(v_1, v_2)
\end{bmatrix} \text{ with } \gamma(v_1, v_2) = \frac{\pi}{4} + 2 \arctan \frac{v_2}{1+v_1}.
\end{equation}
\hfill$\blacksquare$

\section{Proof of Theorem \ref{thm:spline_good1}}\label{sec:spline_good1}

Without loss of generality, we assume that all biases in $f_{\theta}$ are zero as they can be integrated into the $\sigma_\ell$.
First, we show that solutions for \eqref{eq:optim_network} exist.
For $\ell=1,\ldots,(L-1)$ and $\bar \sigma_{1} = \sigma_{1}$, we can iteratively replace the $\sigma_\ell$ in $f_{\theta}$ by $u_\ell$ without changing the output as in
 \begin{align}
     \sigma_{\ell + 1}\bigl(\mathbf{W}_{\ell+1} \bar \sigma_{\ell}(\mathbf{z})\bigr) = \sigma_{\ell + 1}\bigl(\mathbf{W}_{\ell+1}( \bar \sigma_{\ell}(\mathbf{z}) - \bar \sigma_{\ell}(\mathbf{0})) + \mathbf{W}_{\ell+1}\bar \sigma_{\ell}(\mathbf{0})\bigr) = \bar{\sigma}_{\ell + 1}\bigl(\mathbf{W}_{\ell} u_{\ell}(\mathbf{z})\bigr),
 \end{align}
where $u_{\ell} = \bar \sigma_{\ell} - \bar \sigma_{\ell}(\mathbf{0})\in \operatorname{BV}^{(2)}(\R)$ and $\bar{\sigma}_{\ell + 1} = \sigma_{\ell + 1}(\,\cdot + \mathbf{W}_{\ell}\bar \sigma_{\ell}(\mathbf{0}))$.
For the last layer, we set
\begin{align}
\bar \sigma_{L}\bigl(\mathbf{W}_{L} u_{L-1}(\mathbf{z})\bigr) = u_{L}\bigl(\mathbf{W}_{L} u_{L-1}(\mathbf{z})\bigr)+ \mathbf{a}_L,
\end{align}
where $u_{L} = \bar \sigma_{L} - \bar \sigma_{L}(\mathbf{0})\in \operatorname{BV}^{(2)}(\R)$ and $\mathbf{a}_L=\bar \sigma_{L}(\mathbf{0}) = f_\theta(\mathbf{0}) \in \R^{N_L}$.
Consequently, $\TV{2}\left(u_{\ell, n}\right)=\TV{2}\left(\sigma_{\ell, n}\right)$ and $u_{\ell}(\mathbf{0})=\mathbf{0}$.
Hence, a solution of \eqref{eq:optim_network} exists if the restricted problem
\begin{equation}\label{eq:optim_network_reformulation}
\begin{aligned}
&\argmin_{\substack{\mathbf{W}_{\ell}, \sigma_{\ell, n} \in \operatorname{BV}^{(2)}(\mathbb{R})\\\text{s.t.} \operatorname{Lip}(\sigma_{\ell, n}) \leq 1, \, \|\mathbf{W}_{\ell}\| \leq 1\\
\sigma_{\ell,n}(0)=0,|\sigma_{\ell,n}(1)| \leq 1
\\
\mathbf{a}_L\in{\mathbb{R}}^{N_{L}}}} \left(\sum_{m=1}^{M} E\bigl(\mathbf{y}_{m}, f_{\theta}\left(\mathbf{x}_{m}\right) + \mathbf{a}_L\bigr)+\lambda \sum_{\ell=1}^{L} \sum_{n=1}^{N_{\ell}} \TV{2}\left(\sigma_{\ell, n}\right)\right)
\end{aligned}
\end{equation}
has a nonempty solution set.
Since $f_{\theta}$ is $1$-Lipschitz, it holds that
\begin{equation}
    \|f_{\theta}(\mathbf{x}_{m}) - \mathbf{a}_L\| = \|f_{\theta}(\mathbf{x}_{m}) - f_{\theta}(\mathbf{0})\| \leq \|\mathbf{x}_m\|.
\end{equation}
In addition, we have that
\begin{equation}
    2\|\mathbf{a}_L\| = \|(f_{\theta}(\mathbf{x}_{m}) - \mathbf{a}_L) - (f_{\theta}(\mathbf{x}_{m}) + \mathbf{a}_L)\| \leq \|f_{\theta}(\mathbf{x}_{m}) - \mathbf{a}_L\| + \|f_{\theta}(\mathbf{x}_{m}) + \mathbf{a}_L\|
\end{equation}
and, therefore, that
\begin{equation}
    \|f_{\theta}(\mathbf{x}_{m}) + \mathbf{a}_L\|\geq 2\|\mathbf{a}_L\| - \|\mathbf{x}_{m}\|.
\end{equation}
The fact that $E$ is coercive and positive implies that
\begin{equation}
    \lim_{\|\mathbf{a}_L\|\to +\infty} \sum_{m=1}^{M} E\bigl(\mathbf{y}_{m}, f_{\theta}\left(\mathbf{x}_{m}\right) + \mathbf{a}_L\bigr) = +\infty.
\end{equation}
We conclude that there exists a constant $A>0$ such that it suffices to minimize over $\|\mathbf{a}_L\|\leq A$ in \eqref{eq:optim_network_reformulation}.
Now, the remaining steps for the proof of existence are the same as in the proof of Theorem~3 in \cite{AGCU2020}.

For the second part of the claim, we follow the reasoning in \cite{Unser2019}.
Let $f_{\tilde{\theta}}$ be a solution of \eqref{eq:optim_network} with weights \smash{$\tilde{\mathbf{W}}_\ell$}, biases \smash{$\tilde{\mathbf{b}}_\ell$}, and activation functions $\tilde{\sigma}_{\ell, n}$.
When evaluating $f_{\tilde{\theta}}$ at the data point $\mathbf{x}_{m}$, we iteratively generate vectors $\mathbf{z}_{m, \ell}, \tilde{\vec y}_{m, \ell} \in \mathbb{R}^{N_{\ell}}$ as follows.

\begin{enumerate}
    \item Initialization (input of the network): $\Tilde{\mathbf{y}}_{m, 0}=\mathbf{x}_{m}$.
    \item Iterative update: For $\ell=1, \ldots, L$, calculate
\begin{equation}
    \mathbf{z}_{m, \ell}=\left(z_{m, \ell, 1}, \ldots, z_{m, \ell, N_\ell}\right)=\tilde{\mathbf{W}}_{\ell} \tilde{\mathbf{y}}_{m, \ell-1} + \tilde{\mathbf{b}}_{\ell}
\end{equation}
and define $\tilde{\mathbf{y}}_{m, \ell}=\left(\tilde{y}_{m, \ell, 1}, \ldots, \tilde{y}_{m, \ell, N_\ell}\right) \in \mathbb{R}^{N_{\ell}}$ as
\begin{equation}
\tilde{y}_{m, \ell, n}=\tilde{\sigma}_{\ell, n}\left(z_{m, \ell, n}\right), \quad n=1, \ldots, N_{\ell}.
\end{equation}
\end{enumerate}
We directly observe that $\Tilde{\mathbf{y}}_{m,\ell}$ only depends on the values of $\tilde{\sigma}_{\ell, n}\colon \mathbb{R} \rightarrow \mathbb{R}$ at the locations $z_{m, \ell,n}$.
Hence, the optimal $\tilde{\sigma}_{\ell, n}$ are the $1$-Lipschitz interpolations between these points with minimal second-order total variation, as the regularizers for $\tilde{\sigma}_{\ell, n}$ in \eqref{eq:optim_network_reformulation} do not depend on each other.
More precisely, the $\tilde{\sigma}_{\ell, n}$ solve the problem
\begin{equation}\label{eq:var_lip_1d}
    \tilde{\sigma}_{\ell, n} \in \argmin _{\substack{f \in \mathrm{BV}^{(2)}(\mathbb{R})\\  \text{s.t.}  \operatorname{Lip}(f) \leq 1}}\TV{2}(f) \quad\text{s.t.} \quad f\left(z_{m, \ell,n}\right) = \Tilde{y}_{ m, \ell,n},\; m=1, \ldots, M.
\end{equation}
We assume that ${z}_{ m, \ell,n}$ are distinct for $m=1, \ldots, M$.
Otherwise, we can remove the duplicates as ${z}_{ m_1, \ell,n}={z}_{ m_2, \ell,n}$ implies that $\Tilde{y}_{ m_1, \ell,n}=\Tilde{y}_{ m_2, \ell,n}$.
The unconstrained problem
\begin{equation}\label{eq:var_lip_1d_2}
    \min _{\substack{f \in \mathrm{BV}^{(2)}(\mathbb{R})}}\TV{2}(f) \quad\text{s.t.} \quad f\left(z_{m, \ell,n}\right) = \Tilde{y}_{ m, \ell,n},\; m=1, \ldots, M,
\end{equation}
has a linear-spline solution $v_{\ell,n}$ with no more than $M-2$ knots \citep[Proposition 5]{DEBARRE2022114044}.
It can be shown \citep{aziznejad_sparsest_2021} that the Lipschitz constant of this \emph{canonical solution} is given by $\max_{m_1\neq m_2}|\Tilde{y}_{ m_1, \ell,n}-\Tilde{y}_{ m_2, \ell,n}|/|{z}_{ m_1, \ell,n}-{z}_{ m_2, \ell,n}|\leq \mathrm{Lip}(\tilde{\sigma}_{\ell, n})$. Hence, there exists a linear-spline $v_{\ell,n}$ with $\tilde{\sigma}_{\ell, n}({z}_{ m, \ell,n}) = v_{\ell,n}({z}_{ m, \ell,n})$ for all $m=1, \ldots, M$, $\Lip(v_{\ell,n}) \leq \Lip(\tilde{\sigma}_{\ell, n})$, and $\TV{2}(v_{\ell,n}) = \TV{2}(\tilde{\sigma}_{\ell, n})$.
\hfill$\blacksquare$

\section{Properties of SplineProj }\label{sec:splineproj_app}

\paragraph{The Least-Square Projection onto $\{\vec c \in \mathbb{R}^K: \|\mat D\vec c\|_{\infty} \leq T\}$ Preserves the Mean:} Let $\vec x \in \mathbb{R}^K$ and $\vec y \in \{\mathbf{c} \in \mathbb{R}^K: \|\mathbf{Dc}\|_{\infty} \leq T\}$ and $\vec x = \bar{\vec x} + \mu_x \vec 1$, $\vec y = \bar{\vec y} + \mu_y \vec 1$, where $\bar{\vec x}$ and $\bar{\vec y}$ have zero mean.
It holds that
\begin{equation}
    \|\vec x - \vec y\|_2^2 = \|\bar{\vec x} - \bar{\vec y} + \vec 1 (\mu_x - \mu_y)\|_2^2 = \|\bar{\vec x} - \bar{\vec y}\|_2^2 + (\mu_x - \mu_y)^2 K^2.
\end{equation}
Hence, we can add $(\mu_x - \mu_y) \vec 1$ to $\vec y$ and decrease $\|\vec x - \vec y\|$ without violating $ \|\mat D\vec c\|_{\infty} \leq T$.

\paragraph{SplineProj Maps $\mathbb{R}^K$ to $\{\vec x \in \mathbb{R}^K: \|\mat D\vec x\|_\infty \leq T\}$:} We have, for any $\vec c \in \mathbb{R}^K$, that
\begin{equation}
\begin{aligned}
\|\mat D \operatorname{SplineProj}(\vec c)\|_\infty & = \|\mat D \mat D^\dagger \operatorname{Clip}_T(\mat D \vec c) + \mat D \vec 1 \frac{1}{K}\sum_{k=1}^{K}c_k\|_\infty  = \|\operatorname{Clip}_T(\mat D\vec c)\|_\infty \leq T.
\end{aligned}
\end{equation}
Here, we used the fact that $\mat D \mat D^\dagger = \textbf{Id} \in \mathbb{R}^{K-1,K-1}$ and that $\mat D \vec 1 = \mathbf{0} \in \mathbb{R}^K$.

\paragraph{SplineProj is a Projection:} Using the same properties as above, it holds that
\begin{align}
    \operatorname{SplineProj}(\operatorname{SplineProj(\vec c)})& = \mat D^\dagger \operatorname{Clip}_T(\mat D \mat D^\dagger \operatorname{Clip}_T(\mat D \vec c) + \mat D \vec 1 \frac{1}{K} \sum_{k=1}^{K} c_k) + \vec 1 \frac{1}{K} \sum_{k=1}^{K} c_k \notag \\
    & = \mat D^\dagger \operatorname{Clip}_T(\operatorname{Clip}_T(\mat D \vec c)) + \vec 1 \frac{1}{K}\sum_{k=1}^{K}c_k\notag \\
    & = \mat D^\dagger \operatorname{Clip}_T(\mat D \vec c) + \vec 1 \frac{1}{K}\sum_{k=1}^{K} c_k = \operatorname{SplineProj}(\vec c).
\end{align}

\paragraph{SplineProj Preserves the Mean of $\vec c$:} From the properties of the Moore-Penrose inverse, we have that $\operatorname{ker}((\mat D^\dagger)^T) = \operatorname{ker}\left(\mat D\right)$, therefore, $\vec 1^T \mat D^\dagger = \vec 0$ and

\begin{equation}
\begin{aligned}
   \frac{1}{K} \vec 1^T \operatorname{SplineProj}(\vec c) &= \frac{1}{K} \vec 1^T \mat D^\dagger \operatorname{Clip}_T(\mat D \vec c) + \vec 1^T \vec 1 \frac{1}{K^2} \sum_{k=1}^{K} c_k = \frac{1}{K} \sum_{k=1}^{K} c_k.
\end{aligned}
\end{equation}

\paragraph{SplineProj is Differentiable Almost Everywhere with Respect to $\vec c$:} 
The $\operatorname{Clip}_T$ function is differentiable everywhere except at $T$ and $-T$.
Therefore, $\mat D^\dagger \operatorname{Clip}_T(\mat D \vec c)$ is differentiable everywhere except on
\begin{equation}
S = \bigcup_{k=1}^{K-1} \bigl\{\vec x \in \mathbb{R}^K \colon |(\mat D\vec x)_k| = T\bigr\}.
\end{equation}
As a union of $2(K-1)$ hyperplanes of dimension $K-1$, the set $S$ has measure zero in $\mathbb{R}^K$.

\section{Second-Order Total Variation}\label{sec:tv2}
Here, we mainly follow the exposition from \cite{Unser2019}. %and a more general approach can be found in \cite{NeuUns2022}.
Let $\mathcal{S}(\mathbb{R})$ denote the Schwartz space of smooth and rapidly decaying test functions equipped with the usual Schwartz topology, see \cite{Schwartz1966}.
The topological dual space of distributions is denoted by $\mathcal{S}^\prime(\mathbb{R})$ and can be equipped with the total-variation norm
\begin{equation}
\|f\|_{\mathcal{M}} \coloneqq \sup_{\varphi \in \mathcal{S}(\mathbb{R}): \|\varphi\|_{\infty} \leq 1} \langle f, \varphi \rangle.
\end{equation}
As $\mathcal{S}(\mathbb{R})$ is dense in $C_0(\mathbb{R})$, the associated space $\mathcal{M}(\mathbb{R}) = \{f \in \mathcal{S}^\prime(\mathbb{R}): \Vert f \Vert_{\mathcal M} < \infty\}$ can be identified as the Banach space of Radon measures.
Next, we require the second-order distributional derivative $\mathrm{D}^2\colon \mathcal{S}^\prime(\mathbb{R}) \to \mathcal{S}^\prime(\mathbb{R})$ defined via the identity
\begin{equation}   \label{eq:TVmeasure}
   \langle \mathrm{D}^2 f, \varphi \rangle = \Bigl\langle f, \frac{\dx^2}{\dx x^2} \varphi \Bigr\rangle \qquad \forall \varphi \in \mathcal S(\R).
\end{equation}
Based on this operator, the second-order total variation of $f \in \mathcal{S}^\prime(\mathbb{R})$ is defined as
\begin{equation}
\TV{2}(f) = \Vert \mathrm{D}^2 f \Vert_{\mathcal M} = \sup_{\varphi \in \mathcal{S}(\mathbb{R}): \|\varphi\|_{\infty} \leq 1} \Bigl\langle f, \frac{\dx^2}{\dx x^2} \varphi \Bigr\rangle.
\end{equation}
In order to make things more interpretable, we introduce the space of continuous functions $C_{b,1}(\R) = \{f \in C(\R): \Vert f \Vert_{\infty,1} < \infty\}$ that grow at most linearly, which is equipped with the norm $\Vert f \Vert_{\infty,1} \coloneqq \sup_{x \in \R} \vert f(x) \vert (1 + \vert x \vert)^{-1}$.
Now, we are ready to define the space of distributions with bounded second-order total variation as
\begin{equation}
\BV{2}(\mathbb{R}) = \{f \in \mathcal{S}^\prime(\R): \TV{2}(f) < \infty \} = \{f \in C_{b,1}(\R): \TV{2}(f) < \infty \}.
\end{equation}
Note that $\TV{2}(f) = 0$ if $f$ is affine and, consequently, $\TV{2}$ is only a seminorm.

\paragraph{Scale Invariance of $\TV{2}$:} For $\alpha \neq 0$ and $\sigma\in \BV{2}(\R)$, it holds that the rescaling $\tilde \sigma = \frac{1}{\alpha}\sigma(\alpha \cdot)\in \BV{2}(\R)$ satisfies the following invariance
\begin{align}
\TV{2}(\tilde \sigma) & = \!\sup_{\varphi \in \mathcal{S}(\mathbb{R}): \|\varphi\|_{\infty} \leq 1} \int_{\R} \frac{1}{\alpha}\sigma(\alpha x) \frac{\dx^2}{\dx x^2} \varphi (x) \dx x = \!\sup_{\varphi \in \mathcal{S}(\mathbb{R}): \|\varphi\|_{\infty} \leq 1} \int_{\R} \frac{1}{\alpha^2}\sigma(x) \frac{\dx^2}{\dx x^2} \varphi (x/\alpha) \dx x \notag\\
& = \!\sup_{\varphi \in \mathcal{S}(\mathbb{R}): \|\varphi\|_{\infty} \leq 1} \int_{\R} \sigma(x) \frac{\dx^2}{\dx x^2} \varphi (\cdot/\alpha) (x) \dx x = \TV{2}(\sigma).
\end{align}

\section{Proofs for the PnP Stability Results}\label{sec:pnp_results}

\paragraph{Proposition \ref{prop:stability1}:} If $D$ is $\beta$-averaged with $\beta \leq 1/2$, then $2D - \operatorname{Id}$ is 1-Lipschitz since

\begin{align}
\|(2D-\operatorname{Id})(\mathbf{z}_1 - \mathbf{z}_2)\| & = \|2\beta(R(\vec z_1) - R(\vec z_2)) + (1-2\beta)(\vec z_1 - \vec z_2)\| \notag\\
& \leq 2\beta \|R(\vec z_1) - R(\vec z_2)\| + (1-2\beta)\|\vec z_1 - \vec z_2\| \notag \\
& \leq \|\vec z_1 -\vec z_2\|, \quad \forall \vec z_1, \vec z_2 \in \mathbb{R}^n.
\end{align}
Using this property, we get that
\begin{align}
\|(2D-\operatorname{Id})(\vec x_1^* - \alpha \pmb{\nabla} f(\vec H \vec x_1^*, \vec y_1)) - (2D-\operatorname{Id})(\vec x_2^* - \alpha \pmb{\nabla} f(\vec H \vec x_2^*, \vec y_2))\| \notag \\
\leq \|(\vec x_1^* - \alpha \pmb{\nabla}f(\vec H \vec x_1^*, \vec y_1)) - (\vec x_2^* - \alpha \pmb{\nabla} f(\vec H \vec x_2^*, \vec y_2))\|.
\end{align}
From the fixed-point property of  $\vec x_1^*$ and $\vec x_2^*$, we further get that
\begin{align}
\|2(\vec x_1^* - \vec x_2^*) - (\vec x_1^* - \alpha \pmb{\nabla} f(\vec H \vec x_1^*, \vec y_1)) + (\vec x_2^* - \alpha \pmb{\nabla} f(\vec H \vec x_2^*, \vec y_2))\| \notag \\
\leq \|(\vec x_1^* - \alpha \pmb{\nabla} f(\vec H \vec x_1^*, \vec y_1)) - (\vec x_2^* - \alpha \pmb{\nabla} f(\vec H \vec x_2^*, \vec y_2))\|.
\end{align}
Using the fact that $\pmb{\nabla} f(\vec H \vec x, \vec y) = \vec H^T(\vec H \vec x - \vec y)$ and developing on both sides, we get that
\begin{align}
\langle \vec x_1^* - \vec x_2^*, \vec H^T(\vec H \vec x_2^* - \vec y_2) - \vec H^T(\vec H \vec x_1^* - \vec y_1)\rangle \geq 0.
\end{align}
The claim follows by moving $\vec H^T$ to the other side and using the Cauchy-Schwartz inequality
\begin{align}
\|\vec H (\vec x_1^* - \vec x_2^*)\| \|\vec y_1 - \vec y_2\| \geq \langle \vec H(\vec x_1^* - \vec x_2^*), \vec y_1 - \vec y_2 \rangle \geq \|\vec H(\vec x_1^* - \vec x_2^*)\|^2.
\end{align}

\paragraph{Proposition \ref{prop:stability2}:} We show the relation between the difference of the $k$th iterate of the PnP algorithm and the difference of its starting points using the fact that the matrix $\vec I - \alpha \vec H^T \vec H$ has a spectral norm of one when $\alpha$ has an appropriate value. The modulus is

\begin{align}
\|\vec x_1^k - \vec x_2^k\| & = \|D(\vec x_1^{k-1} - \alpha \vec H^T(\vec H \vec x_1^{k-1} - \vec y_1) - D(\vec x_2^{k-1} - \alpha \vec H^T(\vec H \vec x_2^{k-1} - \vec y_2)\| \notag \\
& \leq K \|(\vec I - \alpha \vec H^T \vec H)(\vec x_1^{k-1} - \vec x_2^{k-1}) - \alpha \vec H^T(\vec y_1 - \vec y_2)\| \notag \\
& \leq K \|\vec x_1^{k-1} - \vec x_2^{k-1}\| + \alpha K \|\vec H\|\|\vec y_1 - \vec y_2\| \notag \\
& \leq K^2\|\vec x_1^{k-2} - \vec x_2^{k-2}\| + \alpha \|\vec H\|(K + K^2) \|\vec y_1 - \vec y_2\| \notag \\
& \leq K^k \|\vec x_1^0 - \vec x_2^0\| + \alpha \|\vec H\| \|\vec y_1 - \vec y_2\| \sum_{n=1}^k K^n.
\end{align}
Taking the limit $k \rightarrow \infty$, we get that $\| \vec x_1^* - \vec x_2^*\| \leq \frac{\alpha \|\vec H\|K}{1 - K} \|\vec y_1 - \vec y_2\|$.
\end{document}